\documentclass[journal]{IEEEtran}

\usepackage{graphicx}
\usepackage{wrapfig}
\usepackage{amsmath}
\usepackage{subfig}
\usepackage{esvect}
\usepackage[utf8]{inputenc}
\usepackage{color}
\usepackage{cite}
\usepackage{amssymb}

\usepackage{empheq} 
\usepackage{amsmath}
\usepackage{commath}
\usepackage{indentfirst}
\usepackage{amsfonts}
\usepackage{xcolor}
\usepackage{gensymb}
\usepackage{float}
\usepackage{dblfloatfix}
\usepackage{multirow}
\usepackage{shortcut}

\DeclareMathOperator*{\argmin}{argmin}
\usepackage[shortlabels]{enumitem}

   \usepackage{tabu}
\begin{document}

\title{Making Invisible Visible: Data-Driven Seismic Inversion with Spatio-temporally Constrained Data Augmentation}

\author{Yuxin Yang$^{\dagger, \diamond}$,  Xitong Zhang$^{\dagger, \triangle}$, Qiang Guan$^{\diamond}$,  and \vspace{\baselineskip} Youzuo Lin$^{\dagger, \star}$
\\

\thanks{$\dagger$:~Earth and Environmental Sciences Division, Los Alamos National Laboratory}
\thanks{$\diamond$:~Department of Computer Science, Kent State University}
\thanks{$\triangle$:~Department of Computational Mathematics, Science and Engineering, Michigan State University}
\thanks{
$\star$~Correspondence to: Y. Lin, ylin@lanl.gov.}

}

\markboth{IEEE Transactions on Geoscience and Remote Sensing}
{Shell \MakeLowercase{\textit{et al.}}: Bare Demo of IEEEtran.cls for IEEE Journals}

\maketitle

\begin{abstract}
Deep learning and data-driven approaches have shown great potential in scientific domains. The promise of data-driven techniques relies on the availability of a large volume of high-quality training datasets. Due to the high cost of obtaining data through expensive physical experiments, instruments, and simulations, data augmentation techniques for scientific applications have emerged as a new direction for obtaining scientific data recently. However, existing data augmentation techniques originating from computer vision, yield physically unacceptable data samples that are not helpful for the domain problems that we are interested in. In this paper, we develop new data augmentation techniques based on convolutional neural networks. Specifically, our generative models leverage different physics knowledge (such as governing equations, observable perception, and physics phenomena) to improve the quality of the synthetic data. To validate the effectiveness of our data augmentation techniques, we apply them to solve a subsurface seismic full-waveform inversion using simulated CO$_2$ leakage data. Our interest is to invert for subsurface velocity models associated with very small CO$_2$ leakage. We validate the performance of our methods using comprehensive numerical tests. Via comparison and analysis, we show that data-driven seismic imaging can be significantly enhanced by using our data augmentation techniques. Particularly, the imaging quality has been improved by 15\% in test scenarios of general-sized leakage and 17\% in small-sized leakage when using an augmented training set obtained with our techniques.
\end{abstract}


\IEEEpeerreviewmaketitle

\section{Introduction}

\IEEEPARstart{S}{eismic} full-waveform inversion~(FWI) has been utilized as a popular tool to extract geophysical properties of the subsurface media. Seismic FWI is a typical ill-posed inverse problem due to the limited data coverage. Traditional numerical approaches solving FWI are physics-oriented, usually based on first-order~(gradient-based) or second-order~(gradient and Hessian-based) optimization techniques~\cite{Virieux-2009-Overview}. In recent years, with advances in algorithms and computing capability, there has been remarkable progress in solving FWI problems. Deep learning methods have provided new ways to exploit this abundance of data. Particularly, data-driven seismic inversion techniques have been recently developed~\cite{zhang2020data, wu-2019-inversionnet, araya2018deep}. 

Most of the current data-driven seismic FWI techniques are built-on end-to-end neural network structures. In order to improve the inversion accuracy and the model generalization, data-driven techniques are usually trained on a large volume of dataset, which in turn significantly increases the complexity of the networks. In \cite{wu-2019-inversionnet}, an encoder-decoder structure is developed to learn the regression correspondence from raw seismic data to velocity maps. In \cite{araya2018deep}, a fully-connected network structure is designed for the inversion. In \cite{zhang2020data}, a generative model is utilized and trained for learning the inversion operator. To give an idea of the size of the training data, in \cite{wu-2019-inversionnet}, a ten-layer encoder-decoder results in more than 40 million learnable model parameters. To train this deep neural network, more than 60,000 pairs of labeled simulations need to be available as reported in \cite{wu-2019-inversionnet}. However, obtaining such a large amount of data is challenging (or even infeasible) for some subsurface applications due to the practical obstacles in data acquisition and simulation. Particularly, seismic FWI for monitoring is notoriously known as a ``small-data regime''. A limited number of seismic sensors are distributed over a large area, and very few time-lapse observations can be affordably acquired~\cite{Lumley-2001-Time}. Training a data-driven seismic FWI using limited data will result in weak generalizability and misfitting. To fully unleash the power of deep learning for a better, faster, and cheaper subsurface seismic FWI approach, we develop a new data augmentation technique to bridge the gap by addressing the critical issues of generalizability and the capability of generating high quality and a large volume of training data. 

Data augmentation, the process of creating new samples by manipulating the original data, addresses the data shortage at the root of the problem~\cite{Shorten-2019-survey}. However, the most popular data augmentation methods are not appropriate for seismic imaging due to their inability to incorporate generic physics properties. Furthermore, seismic data usually yields both spatial and temporal characteristics. To address those issues, we develop data augmentation techniques to account for both spatio-temporal characteristics and the critical seismic physics to generate high-quality simulations. Our models are built on variation autoencoder~(VAE) to take advantage of its direct tie to the latent representations~\cite{VAE-2014-Diederik}. The design of  our techniques considers different representations of physics including the governing equations, the observable perception, and the physics phenomena. To validate the performance of our developed techniques, we test our models using an existing CO$_2$ leakage synthetic dataset, Kimberlina dataset, generated and operated by the U.S. Department of Energy~(DOE)~\cite{Characterizing-2017-DOE}. Our interest is to employ our data-driven FWI to image and detect small CO$_2$ leaks. Via various numerical tests, we demonstrate that our data augmentation techniques significantly improve the data representativeness of the training set, which in turn enhances the seismic imaging accuracy. Specifically, CO$_2$ plumes related to small leaks can now be much better imaged than those obtained without using augmentation.  

In the following sections, we first briefly provide the related work in Section~\ref{sec:relatedwork}. We then provide the fundamentals of physics-driven versus data-driven methods in networks~(Section~\ref{sec:background}). Besides, we also discuss the data option and the technical challenges of the problem in Section~\ref{sec:background}.  We develop and discuss our  data augmentation techniques in Section~\ref{sec:methods}. We further provide all the numerical tests and results in Section~\ref{sec:results}.  Finally, further discussion, future work, and concluding remarks will be presented.

\section{Related Work}
\label{sec:relatedwork}

\subsection{Deep Generative Models}

Generative models are known as a type of unsupervised learning approaches that explicitly or implicitly model the distribution of true data so as to generate new samples with some variations~\cite{Bishop-2006-Pattern}. Current state-of-the-art generative models are built on deep neural networks (i.e., deep generative models (DGMs)). Examples of recent DGMs include variational autoencoders~(VAE)~\cite{VAE-2014-Diederik} and generative adversarial networks~(GAN)~\cite{goodfellow2014generative}. 

As a variation in autoencoder, VAE belongs to the DGMs that learn the data distribution explicitly. It solves a variational inference problem to maximize the marginalized data likelihood by using a generative network~(decoder) and a recognition network~(encoder). Once fully trained, the encoder learns a distribution over latent variable given observation, and the decoder learns a distribution over observation given latent variable. VAE and its variants have shown great potential in generating data for augmentation in different applications~\cite{Luo-2020-data, hsu2017unsupervised,nishizaki2017data,liu2018data}. In particular, \cite{Luo-2020-data} employ a vanilla VAE to generate synthetic EEG time series for recognizing emotions. \cite{nishizaki2017data} also employ VAE to generate waveform data for automatic speech recognition. In \cite{liu2018data}, VAE is used to extracted useful features in the latent space from image data. Linear interpolation on the latent space is conducted to obtain new synthetic images. \cite{hsu2017unsupervised} develop a VAE-based data augmentation technique to address the distribution mismatch in source and target domains for improving the performance of a domain adaptation method in speech recognition. 
Besides VAE, other DGMs (such as GAN) have also been applied for the task of data augmentation~\cite{zhang2019dada,antoniou2017data,shrivastava2017learning,sixt2018rendergan}. In comparison, VAE provides a natural connection to the data distribution by collapsing most dimensions in the latent representations. Another noticeable benefit of VAE-based DGMs is the relatively easier effort to train with less technical complexity for hyper-parameter selection.  Provided with these aforementioned encouraging results of DGMs, a direct application of DGMs to our problems may face two major challenges. Firstly, DGMs are in general highly data-demanding. Secondly, they are purely driven by data without considering physics.

\subsection{Physics-Informed Deep Learning}

Physics-informed~(i.e., domain-aware) learning is a critical task to scientific machine learning~(SciML) community~\cite{Basic-2019-DOE}. Particularly, how to incorporate physics information becomes one of the most challenging and important research topics across different scientific domains~\cite{sun2020surrogate, Gomez-2020-Physics,wang2020towards,  raissi2019physics,zhu2019physics}. A thorough survey on this topic is published by~\cite{Physics-2021-Karniadakis, Willard-2020-Integrating}. As pointed out in~\cite{Physics-2021-Karniadakis}, there are three ways to make a learning algorithm physics-informed, ``observation bias'', ``inductive bias'', and ``learning bias''. The observation bias approaches introduce physics to the model directly through data that embody the underlying knowledge. The inductive bias approaches focus on designing neural network architectures that implicitly enforce physics knowledge associated with a given predictive task. The learning bias approaches incorporate the physics knowledge in a soft manner by appropriately penalizing the loss function of conventional neural networks. Our approach developed in this work belong to two categories of the above: observation bias and learning bias.

There are many benefits considering physics knowledge when designing a neural network models. Regardless of the application domains, one of the major benefits is to improve the robustness of the prediction model and to produce physically meaningful (and more accurate) results. Particularly, \cite{lagaris1998artificial} propose an artificial neural network method to solve partial differential equations~(PDEs) for flow simulations. \cite{raissi2019physics} develop a deep-learning-based nonlinear PDE solver. \cite{zhu2019physics} develop a numerical PDE solver using a convolutional encoder-decoder and a flow-based generative model with physics constraints. More accurate results have been shown in their work. \cite{sun2020surrogate} develop another PDE solver using the physics-informed deep learning method. Their method leverages both the full-physics simulations and additional physics-based constraints. \cite{wang2020towards} develop a spatiotemporal deep learning model to account for both data characteristics and underline physics to synthesize high-quality turbulent imagery. All these aforementioned works provide us with great inspiration about leveraging useful physics information while developing deep learning models for our seismic imaging problems. 
\begin{figure*}[ht]
	\centerline{
	\subfloat[]{\includegraphics[width=.20\linewidth]{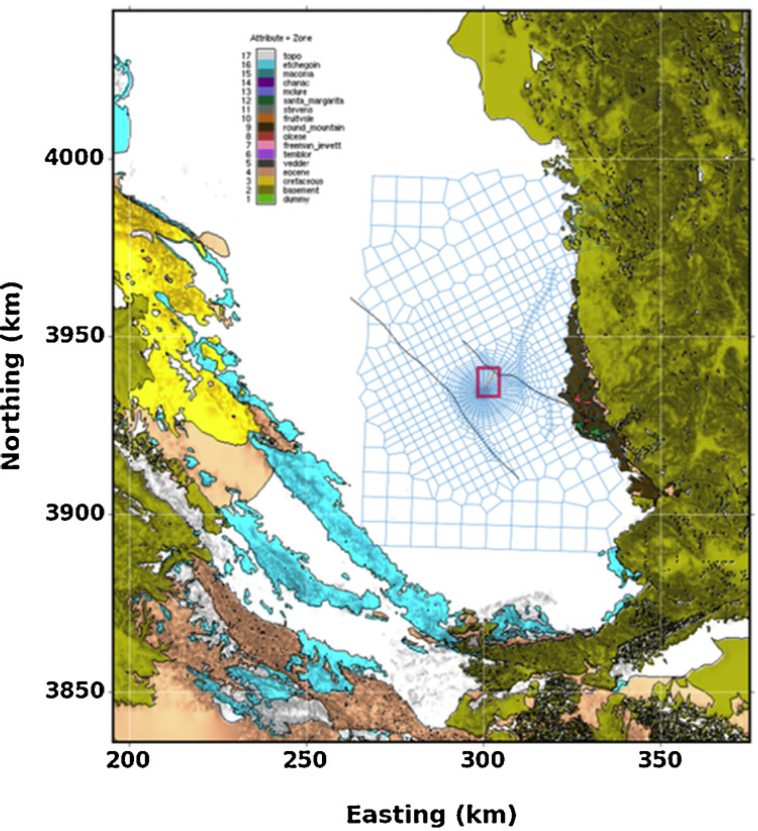}}
	\quad \quad \quad \quad
	\subfloat[]{\includegraphics[width=.13\linewidth]{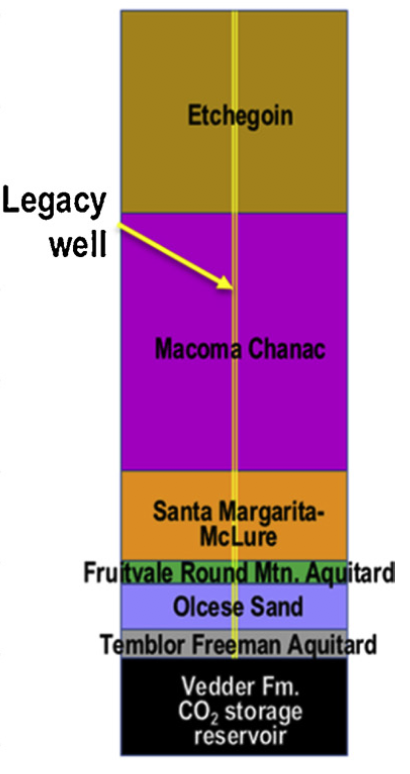}}
	\quad \quad \quad \quad
	\subfloat[]{\includegraphics[width=.28\linewidth]{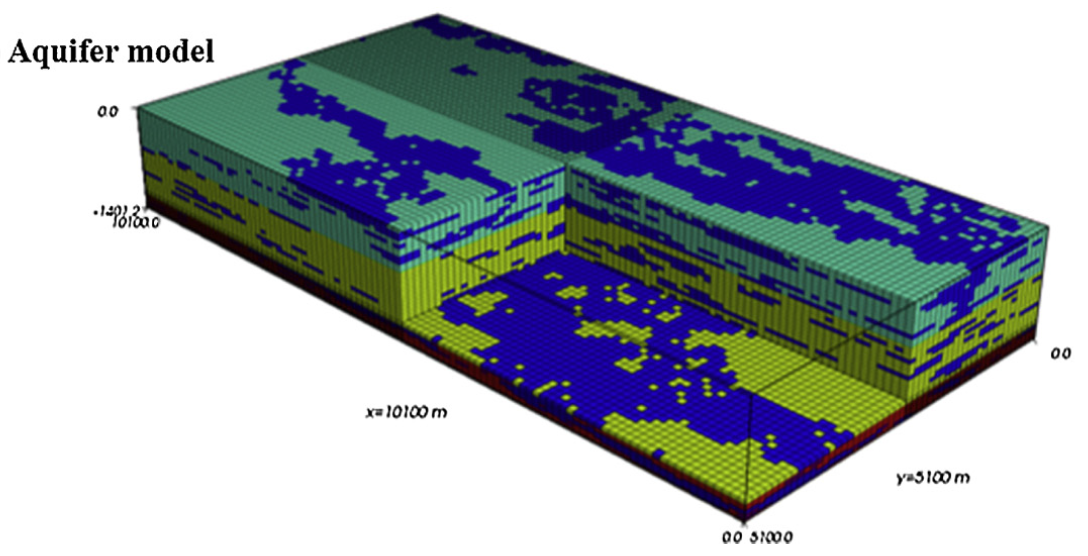}}}
    \vspace{0.5cm}
	\centerline{
	\subfloat[]{\includegraphics[width=\linewidth]{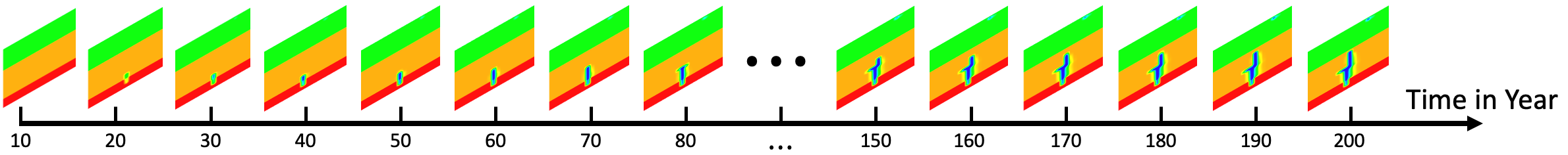}}}
	\caption{Illustration of the Kimberlina dataset and three modeling modules used to generate the simulated velocity maps. (a) CO$_{2}$ storage reservoir model, (b) wellbore leakage model, (c) multi-phase flow and reactive transport models of CO$_{2}$ migration in aquifers~\cite{Downhole-2019-Buscheck, Assessment-2019-Yang}, and (d) Illustration of a set of simulation with 20 velocity maps over a duration of 200 years. A CO$_2$ leakage will result in a decrease of the velocity value in the location where the leak happens.}
	\label{fig:kimberlina}
\end{figure*}

\subsection{Data Augmentation in Seismic Exploration}

In the seismic exploration community, there has been surprisingly little work addressing this dilemma of lack of data for data-driven seismic FWI. The existing approaches can be roughly categorized into two groups, those based on velocity building~\cite{Liu-2021-Deep, Ren-2021-Building, Wu-2020-Building} and those based on pure machine learning approaches~\cite{Physically-2020-Feng, Gomez-2020-Physics, ovcharenko2019style}. Specifically, in \cite{Liu-2021-Deep} and \cite{Ren-2021-Building}, a large volume of subsurface velocity maps are generated to include different geologic structures. The geometry of those pre-generated geologic structures is assumed to follow a certain distribution. \cite{Wu-2020-Building} design a workflow to automatically build a subsurface structure with folding and faulting features. Their method relies on the initial layer-like structure, therefore, producing unsatisfactory results when applying to different sites. In ~\cite{Gomez-2020-Physics}, an adaptive data augmentation technique is developed to augment the training by using unlabeled seismic data. \cite{Physically-2020-Feng} develop a style transfer technique to generate synthetic velocity maps from natural images. \cite{ovcharenko2019style} develop a set of subsurface structure maps using customized subsurface random model generators. Their method strongly relies on domain knowledge to generate the content images, which in turn significantly limits the variability of the training set.

\section{Background}
\label{sec:background}

\subsection{Data-driven Seismic Full-Waveform Inversion}
Seismic acoustic-wave equation in the time-domain can be given by 
\begin{equation}
 \left [ \frac{1}{K(\mathbf{r})} \frac{\partial ^2}{\partial t ^2} 
        - \nabla  \cdot \left ( \frac{1}{\rho (\mathbf{r})}\,\, \nabla 
        \right ) \right ]
        p(\mathbf{r}, t) = s(\mathbf{r},\, t),
\label{eq:Forward}
\end{equation}
where $\rho (\mathbf{r})$ is the density at spatial location $\mathbf{r}$, $K(\mathbf{r})$ is the bulk modulus, $s(\mathbf{r},\, t)$ is the source term, $p(\mathbf{r}, t)$ is the pressure wavefield,  and $t$ represents time. In this work, we use a time-domain finite-difference scheme to solve the wave equation for generating synthetic seismic data~\cite{Finite-2007-Moczo}. This finite-difference scheme has 8$^\text{th}$-order accuracy in space and 2$^\text{th}$-order accuracy in time. 

Due to the ill-posedness nature of full-waveform inversion, it is challenging to solve FWI problem. A new emerging computational approaches solving FWI is based on deep learning. In our recent work, we develop an encoder-decoder-based end-to-end network structure and name it ``InversionNet''~\cite{wu-2019-inversionnet}. In particular, InversionNet learns an inversion operator directly from seismic data to the velocity map. InversionNet consists of a series of convolution blocks with a set of trainable parameters $\theta$. By denoting the target inversion operator as $\mathcal{G}$, and further assuming a training set $\mathcal{D}=\{(\mathbf{x}_{1}, \mathbf{y}_{1}), (\mathbf{x}_{2}, \mathbf{y}_{2}), ..., (\mathbf{x}_{n}, \mathbf{y}_{n})\}$, where $\mathbf{x}$ is seismic data and $\mathbf{y}$ is velocity map, the InversionNet is to obtain $\mathcal{G}$ by solving the following optimization problem
\begin{equation}
    \mathcal{G} = \argmin_{\mathcal{G}} \left \{ \sum_{i=1} ^{n} \| \mathbf{y}_i - \mathcal{G} (\mathbf{x}_i) \|_1 \right \}.
    \label{eq:eq_mae}
\end{equation}
It is worth mentioning that we use the mean-absolute error (MAE) as the loss function to calculate reconstruction error between the ground truth and the predicted velocity map. We backpropagate the reconstruction error from the output of InversionNet to the first layer of InversionNet to update $\theta$ in order to obtain its optimal value.  Full details of the implementation of InversionNet are provided in our recent work~\cite{wu-2019-inversionnet}.

\subsection{Small CO$_2$ Leak Detection and Kimberlina Dataset}
\label{Sec:data}

In geologic carbon sequestration~(GCS), also known as carbon capture and storage~(CCS), developing effective monitoring methods is urgently needed to detect and respond to CO$_2$ leakage. This is particularly important for early detection, which would provide timely warning and intervention before the potential damages to the environment~(such as acidification of groundwater and killing of plant life, contamination of the atmosphere, etc)~\cite{Carbon-2003-Ha}. On the other hand, detecting small CO$_2$ leaks is also technically challenging since it requires high detectability and sufficient spatial resolution of geophysical methods to capture the subtle geologic feature perturbation induced by the leaks. Considering this pressing need, our goal in this work is to assess and further improve the early CO$_2$ leak-detection capabilities of the seismic FWI method.

To our best knowledge, we are unaware of any available field seismic data that fits the scope of our problem of interest. Meanwhile, this lack of data is recognized by the U.S. Department of Energy~(DOE), and to alleviate this problem, given the importance of this application, the DOE, through the National Risk Assessment Partnership (NRAP) project, has generated a set of high fidelity simulations, the Kimberlina dataset, with the aim of providing a standard baseline dataset to understand and assess the effectiveness of various geophysical monitoring techniques for detecting CO$_2$ leakage~\cite{Characterizing-2017-DOE}. The Kimberlina dataset is generated from a hypothetical numerical model built on the geologic structure of a commercial-scale geologic carbon sequestration reservoir at the Kimberlina site in the southern San Joaquin Basin, 30~km northwest of Bakersfield, CA, USA. The simulation procedure consists of four modules: a CO$_{2}$ storage reservoir model (Fig.~\ref{fig:kimberlina}(a)), a wellbore leakage model (Fig.~\ref{fig:kimberlina}(b)), a multi-phase flow and reactive transport models of CO$_{2}$ migration in aquifers (Fig.~\ref{fig:kimberlina}(c)), and a geophysical model. In particular, the P-wave velocity maps used in this work belong to the geophysical model, which is created based on the realistic geologic-layer properties from the GCS site as shown in Fig.~\ref{fig:kimberlina}(b)~\cite{Downhole-2019-Buscheck, Assessment-2019-Yang}. 

\begin{figure}
    \centering
    \includegraphics[width=0.75\linewidth]{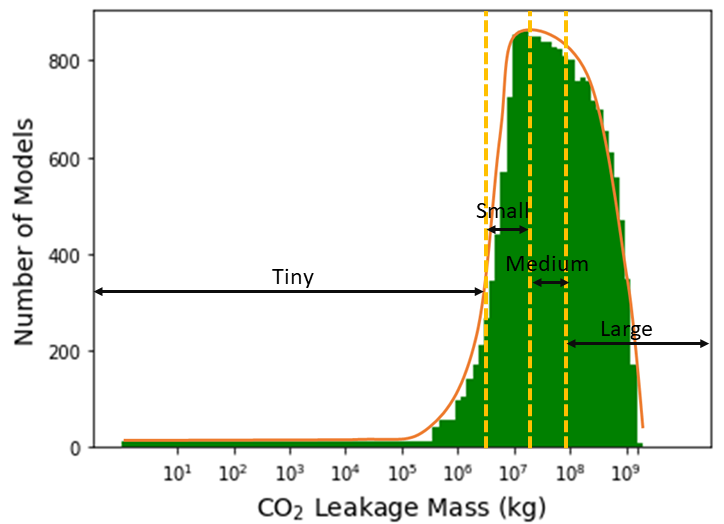}
    \caption{Distribution of leakage mass of Kimberlina Dataset. Each of the splittings covers $20\%$, $20\%$, $20\%$, and $40\%$ of the data samples, respectively. }
    \label{fig:mass dis}
\end{figure}

\begin{figure*}[ht]
	\centerline{
	\subfloat[]{\includegraphics[width=.50\linewidth]{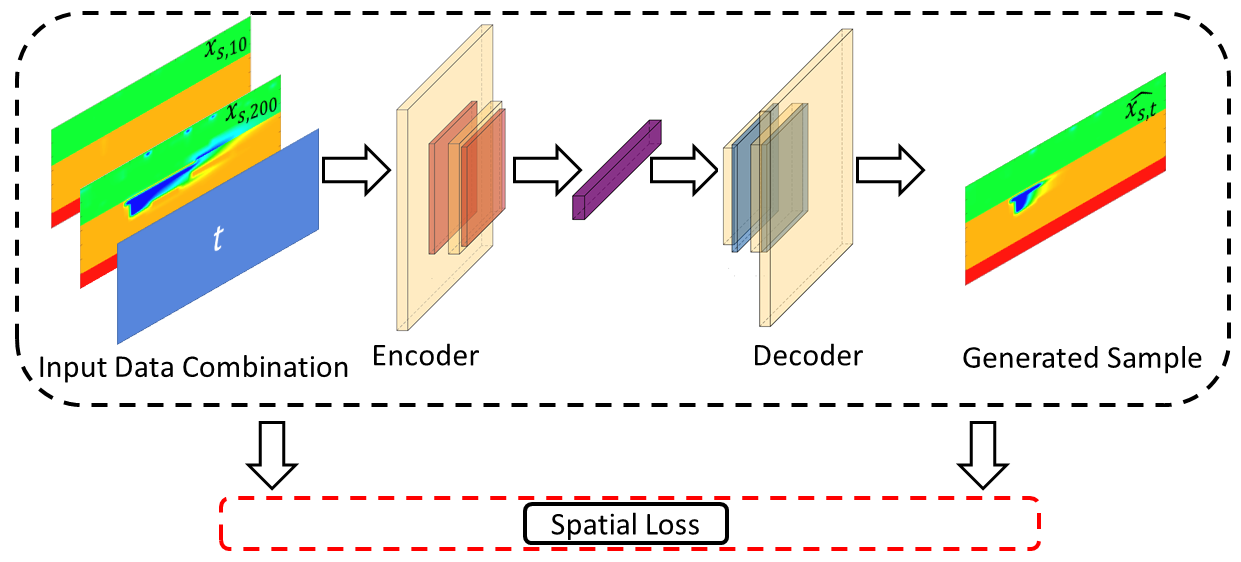}}
	\quad
	\subfloat[]{\includegraphics[width=.50\linewidth]{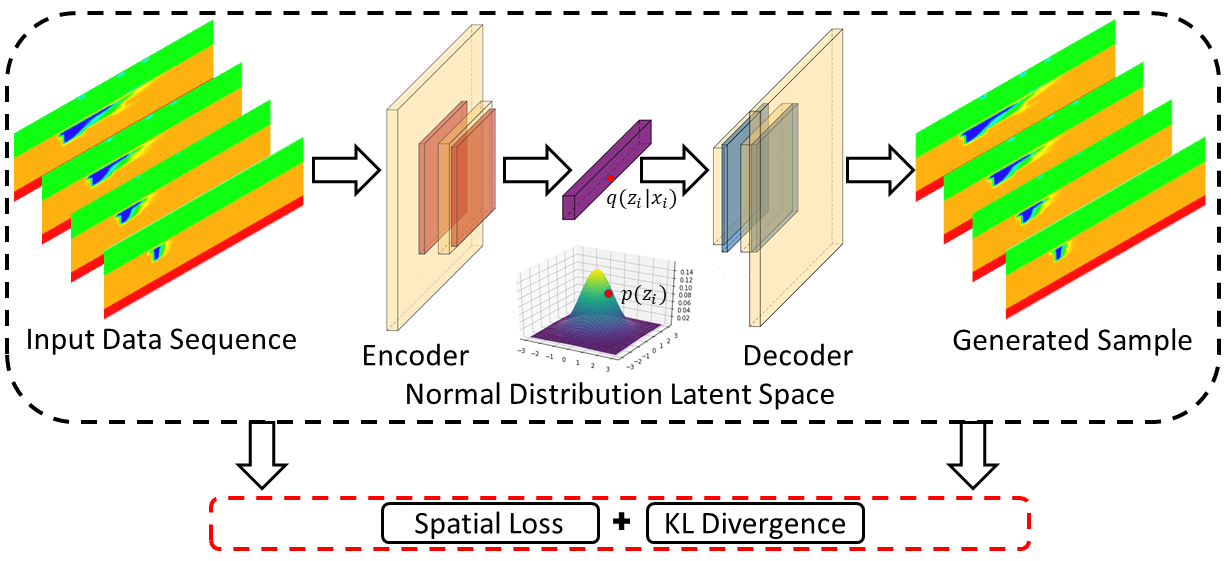}}}
	\caption{Schematic illustration of our (a) autoencoder and (b) VAE generative models.}
	\label{fig:AE_VAE}
\end{figure*}

The Kimberlina dataset contains $991$ CO$_2$ leakage scenarios,
each simulated over a duration of 200 years, with 20 leakage velocity
maps provided~(i.e., at every 10 years) for each scenario. An illustration of one specific leakage simulation associated with the leakage mass over 200 years are shown in Fig.~\ref{fig:kimberlina}(d). We also provide the overall distribution of the whole dataset in Fig.~\ref{fig:mass dis}. For a balanced dataset, we would expect the data label (leakage mass) should be uniformly distributed, which however is not the case for Kimberlina dataset. Particularly, the whole dataset can be split into four parts by its leak mass as
\begin{equation}
    \begin{cases}
      \textbf{\textit{Tiny}} & \text{if \textbf{mass} $\textless~9.10\times 10^{6}\mathrm{~kg}$}, \\
      \textbf{\textit{Small}} & \text{if $9.10\times 10^{6}\mathrm{~kg} \textless$ \textbf{mass}  $\textless 2.67\times 10^{7}\mathrm{~kg}$},\\
      \textbf{\textit{Medium}} & \text{if $2.67\times 10^{7}\mathrm{~kg} \textless$ \textbf{mass}  $\textless 8.05\times 10^{7}\mathrm{~kg}$},\\
      \textbf{\textit{Large}} & \text{if $8.05\times 10^{7}\mathrm{~kg} \textless$ \textbf{mass}}.
    \end{cases}       
    \label{eq:leakCases}        
\end{equation}
Each of the splittings covers $20\%$, $20\%$, $20\%$, and $40\%$ of the data samples, respectively. Although we have $20\%$ of tiny leakage samples, these samples are distributed from $0$ to $9.1\times 10^6$ that covers nearly $70\%$ of CO$_2$ leakage scenarios, as shown in Fig.~\ref{fig:mass dis}. In other words, the density of tiny samples is much lower than that of the other three classes. This sparsity and in-balanced sample density create the major challenge when imaging tiny leakage samples.

The Kimberlina project focuses on the shallow CO$_2$ leakage. That leads to 3-layer synthetic velocity models~(baseline and monitor), which reflect the shallow geologic structure from the field study. The Kimberlina model and simulations have been the basis for a variety of extensive research efforts in characterizing and detecting for CO$_{2}$ using different geophysical approaches~\cite{Time-2020-Appriou, Optimal-2020-Chen, Zheng-2019-Data, Downhole-2019-Buscheck, Assessment-2019-Yang}. Our interest is on the early-leak detection, which requires to image those small leaks.  The unbalancing of the dataset becomes a major challenge for our data-driven seismic inversion technique since it will mislead our InversionNet model towards medium or large leaks. On the other hand, due to the limitation of physical simulations, we will not be able to further generate more synthetic for the small leaks. Hence, those practical obstacles make our problem reside in a low-data regime scenario. Next, we will describe our techniques to augment the Kimberlina dataset while preserving the physics information as much as we can to improve the prediction accuracy of our InversionNet model. 

\section{Methodology}
\label{sec:methods}

\subsection{Physics of the Problem}

Our data augmentation techniques will leverage existing physics knowledge of the problem. It is worth understanding what specific physics information are referred to in this context for the designing and training our neural networks. 
\begin{enumerate}[i.]
\item \textbf{Governing Equations.}~One of the most prominent physics knowledge in our problem is that the governing equations are used to generate original physical simulations~(as shown in Fig.~\ref{fig:kimberlina}). Those equations describe specific physical relationships between time and spatial derivatives explicitly using temporally dynamic formulas. In order to generate physically meaningful synthesized data, it would be important to embed that physics information in the generative models. 

\item \textbf{Observable Perception.}~As described in Section~\ref{Sec:data}, the data that we are interested in synthesizing are two-dimensional~(2D), which means it yields a distribution that would be represented in certain visual perception. We expect our generative model would be able to capture the underline true data distribution, which in turn would require the synthesized 2D data would physically ``look like'' those in the training data.

\item \textbf{Physics Phenomena.}~Any physical simulation should respect the realistic physics phenomena. As one example, in our problem of interest, the super-critical CO$_2$ will migrate over time, meaning that we will observe the spatial spreading of CO$_2$ should gradually increase over time. How to best design our generative model without violating this phenomenon would potentially help to improve the performance of our generative model.  

\end{enumerate}

We will consider all of the above during the development of our  generative models. Another point that would be important to consider is that all of the above physics information is consistent throughout all temporal duration.

\subsection{Data-driven Generative Models}

To compensate for the imbalance data as shown in Fig.~\ref{fig:mass dis}, we would like to generate more data in the small-leak region. Luckily, the original Kimberlina dataset provides full-physics simulation in the medium- and large-leak regions. Those data are generated by the governing physics equations, which means those physics knowledge are represented by those data implicitly. Our first two generative models are built on autoencoder and VAE to leverage those existing simulations while taking into account the temporal variation.

\subsubsection{Autoencoder}

Our first model is to build a ``regression'' model that would provide interpolated data for those temporally missing points~(as shown in Fig.~\ref{fig:AE_VAE}(a)). The hypothesis behind this idea is that considering the consistency of the physics, we would expect that once fully trained, our generative model will capture the intrinsic dynamics of the physics from the existing simulations so that it will provide physically realistic prediction at any given time, particularly, those at the early stage of the leakage. 

\begin{figure*}[ht]
    \centering
    \begin{minipage}[c]{.5\textwidth}
      \centering
      \includegraphics[width=0.80\textwidth]{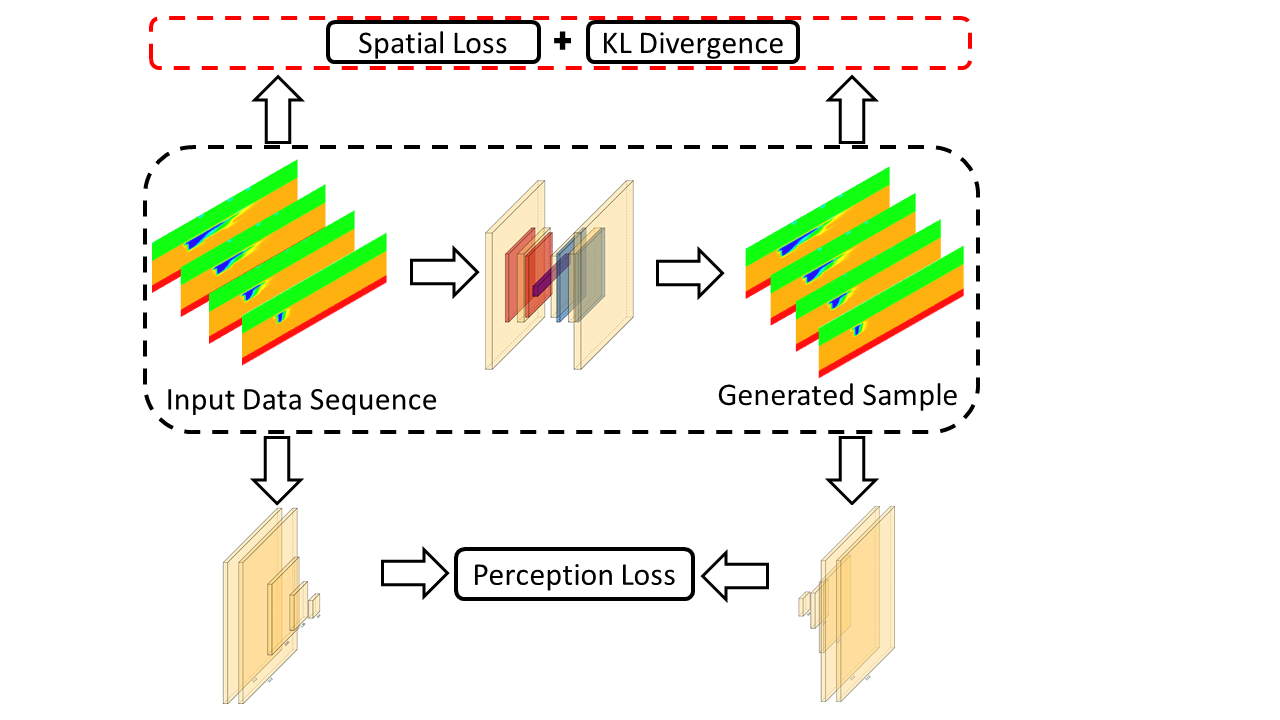}
      \centerline{\small (a)}
    \end{minipage}%
    \begin{minipage}[c]{.5\textwidth}
      \centering
      \includegraphics[width=0.85\textwidth]{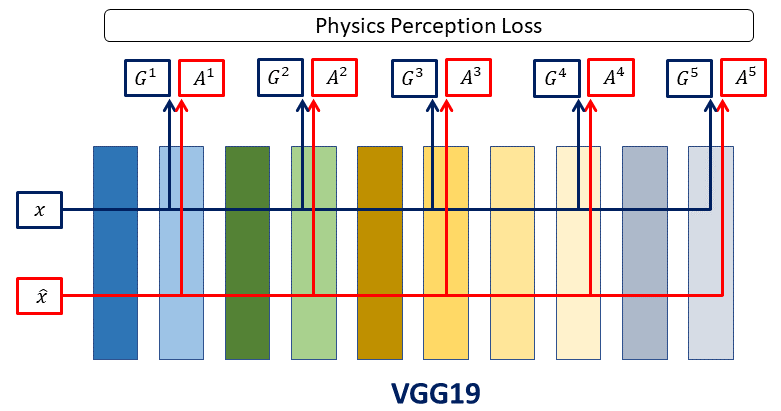}     
      \centerline{\small (b)} 
    \end{minipage}
	\caption{Schematic illustration of (a) our new variational autoencoder with perception loss, and (b) the perception loss using the pre-trained VGG-19 network~\cite{simonyan2014very}.}
	\label{fig:VAE_Style}
\end{figure*}
Technically, our model will be based on an autoencoder structure, which consists of a convolutional encoder $\mathcal{F}_{\theta}$, with a set of trainable parameters $\theta$, and a convolutional decoder $\mathcal{G}_{\phi}$, with a set of trainable parameters $\phi$. To incorporate the spatial information, we set two input channels of our encoder as the first and the last velocity maps from one simulation. To incorporate the temporal information, we further create a temporal matrix by replicating the single time value over all matrix entries. The temporal matrix will then be used as one of the three input channels of the autoencoder together with two other two. When training the autoencoder, all three input channels will be convolved together, leading to the incorporation of both spatial and temporal information. Once fully trained, our encoder will learn to reduce the dimensionality of this mixture of inputs to a latent variable which is a high-level latent representation containing both spatio-temporal information. Our decoder will estimate the target velocity map using the latent variable, which can be considered as nonlinear high-dimensional regression.  

The structure of our generative model is shown as Fig.~\ref{fig:AE_VAE}(a), and mathematically our autoencoder can be represented as 
\begin{equation}
    \begin{split}
    &\mathrm{Encoder:~}z = \mathcal{F}_{\theta}(x_{s, 10},\,x_{s,200},\,t),\\
    &\mathrm{Decoder:~}\hat{x}_{s, t}=\mathcal{G}_{\phi}(z),
    \end{split}
\end{equation}
where $x_{s, 10}$, $x_{s,200}$ and $t$ are the inputs of the encoder. $t$ is the time of the velocity map that needs to be predicted and it is created as a temporal matrix by replicating the single time value over all matrix entries. $x_{s, 10}$ and $x_{s, 200}$ are the first data and the last data from the same simulation $s$, where 10 and 200 are the time index of the data. $z$ is the latent variable output by the encoder $\mathcal{F}$, and the decoder $\mathcal{G}$ produces $\hat{x}_{s, t}$, the estimated velocity map of simulation $s$ at time $t$. We use Mean Squared Error~(MSE) as our optimality criterion to compute the reconstruction loss between the ground truth and the generated velocity maps and to update trainable parameters through backpropagation
\begin{equation}
\begin{split}
    \mathcal{L}(\theta,\phi)&=\mathcal{L}_{recon}=\frac{1}{|S||T|}\sum_{s\in S,t\in T}( x_{s, t} - \hat{x}_{s, t})^2,\\
    &=\frac{1}{|S||T|}\sum_{s\in S,t\in T}(x_{s,t}-\mathcal{G}_{\phi}(\mathcal{F}_{\theta}(x_{s,10},x_{s,200},t)))^2.
\end{split}
    \label{eq:AE_Loss}
\end{equation}

This model generates synthetic samples in the data domain. As discussed in~\cite{Oring-2020-Autoencoder}, generating samples in the latent space might increase the variability within the data distribution. We, therefore, study VAE and its capability in generating samples. 

\subsubsection{Variational Autoencoder}

The variational autoencoder~(VAE) is a probabilistic generative model to create a latent representation of the input data. That would allow us to generate new samples with high diversity by manipulating the latent representations. Unlike the autoencoder model, which incorporates temporal information as part of the input~(Fig.~\ref{fig:AE_VAE}(a)), our VAE generative model produces new temporal interpolation separately in two steps. In the first step, we train the VAE by taking only velocity maps as input without explicit temporal information. Once fully trained, the VAE will be able to generate latent variables representing the velocity maps. In the second step, we provide a linear interpolation scheme on the normal distributed latent space to produce new latent variables for further synthesizing new velocity samples. The idea behind the VAE generative model is somewhat similar to that of the autoencoder. We expect the physics knowledge, i.e., the governing physics relationship can be captured by training the VAE using simulations. The consistency of the physics information will be leveraged when generating new samples at different times.  

We provide the illustration of our VAE generative model in Fig.~\ref{fig:AE_VAE}(b). The encoder, $\mathcal{F}$, and decoder, $\mathcal{G}$, structures of VAE are similar to those of the autoencoder. However, the encoder in VAE is to learn the posterior distribution $q_{\theta}(z|x)$, which is the distribution parameter of latent variable $z$ given input $x$. The decoder in VAE is to learn the conditional distribution $p_{\phi}(x|z)$, which is the distribution of reconstructed data given latent distribution. There is a prior distribution $p(z)$ over the latent space, which we set as a standard normal distribution. The output of encoder $q_{\theta}(z|x)$ has two parts of mean and log-variance of the posterior distribution. One of the known problems associated with VAE is that its gradients cannot flow through the bottleneck of mean and log-variance. So, we perform a re-parameterize trick to make the gradient able to flow through the bottleneck~\cite{VAE-2014-Diederik},
\begin{equation}
    z = \mu + \sigma\odot\epsilon,
    \label{eq:reparameterize_VAE}
\end{equation}
where $z\in\mathbb{R}^{64}$ is the latent sample, $\mu\in\mathbb{R}^{64}$ and $\sigma\in\mathbb{R}^{64}$ are the mean and log-variance of the posterior distribution $q_{\theta}(z|x)$, and $\epsilon\in\mathbb{R}^{64}$ is a random variable sampled from normal distribution and independent from $\mu$ and $\sigma$. We employ the standard VAE loss function as below
\begin{equation}
\begin{split}
    \mathcal{L}(\theta,\phi) &= \mathcal{L}_{recon} + \mathcal{L}_{kld},\\
    &=\sum_{i}\mathbb{E}_{q_{\theta}(z_{i}|x_{i})}\left[\log\frac{p_{\phi}(x_{i},z_{i})}{q_{\theta}(z_{i}|x_{i})}\right],\\
    &=\sum_{i}\mathbb{E}_{q_{\theta}(z_i|x_i)}(\log p_{\phi}(x_i|z_i)\\
    &+\log p(z_i)-\log q_{\theta}(z_i|x_i)),\\
    &=\sum_{i}(x_{i}-\hat{x}_{i})^{2}+\sum_{i}D_{KL}(q_{\theta}(z_{i}|x_{i})\|p(z_{i})),
\end{split}
    \label{eq:VAE}
\end{equation}
where $\mathcal{L}_{kld}=\sum_{i}D_{KL}(q_{\theta}(z|x_{i})\|p(z))=\sum_{i}\mathbb{E}_{q_{\theta}(z_{i}|x_{i})}(\log p(z_i)-\log q_{\theta}(z_i|x_i))$ is to measure the KL-divergence between the posterior distribution $q_{\theta}(z|x)$ and the prior distribution $p(z)$. $\mathcal{L}_{recon}=\sum_{i}\mathbb{E}_{q_{\theta}(z|x_{i})}(\log p_{\phi}(x_i|z_i)=\sum_{i}(x_{i}-\hat{x}_{i})^{2}$ is the reconstruction loss between ground truth velocity maps, $x_i$ and generated velocity maps, $\hat{x}_i$. 

When generating new velocity maps to augment our dataset, a directly random sampling on the prior distribution may lead to velocity maps associated with different leaks, whereas our interest is to obtain more small-leakage velocity maps. So we come up with an interpolation strategy. Particularly, we obtain the latent variables of two adjacent velocity maps from the same simulation, namely, $z_1$ and $z_2$ through the encoder, $\mathcal{F}$. Multiple new latent variables can be  interpolated between these $z_1$ and $z_2$ before passing them through the decoder, $\mathcal{G}$, to further generate additional velocity maps~\cite{berthelot-2018-understanding}. The procedure can be posed as follows
\begin{equation}
\begin{split}
    \hat{x}_{\alpha} &= \mathcal{G}_{\phi}(\alpha z_{1} + (1-\alpha) z_{2}),\\
    &=\mathcal{G}_{\phi}(\alpha\mathcal{F}_{\theta}(x_{s,10})+(1-\alpha)\mathcal{F}_{\theta}(x_{s,200}))
    ,
    \label{eq:VAE_interpolation}
\end{split}
\end{equation}
where $\alpha\in[0,1]$ is the coefficient of the interpolation. Different from the autoencoder model, the temporal information is not used as the input in this VAE generative model.

It is worth mentioning that for either the autoencoder or VAE as shown in Fig.~\ref{fig:AE_VAE}, we expect that the governing physics will be able to be learned through training the models using full-physics simulations. However, other physics knowledge~(such as observable perception or physics phenomena) will not be able to be captured by the generative models. Hence, we come up with two different strategies to further constrain our generative models with additional physics knowledge. 

\begin{figure*}[h]
    \centering
    \begin{minipage}[c]{.5\textwidth}
      \centering
      \includegraphics[width=0.80\textwidth]{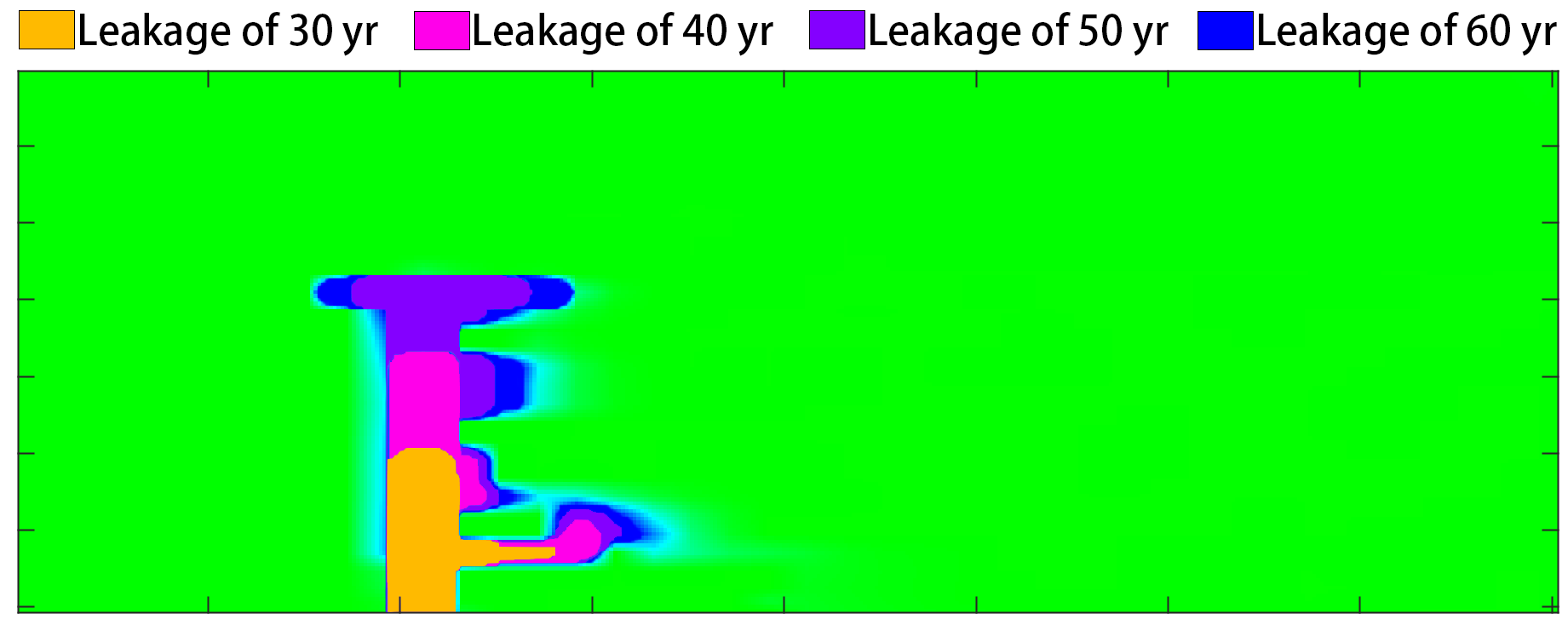}
      \centerline{\small (a)}
    \end{minipage}%
    \begin{minipage}[c]{.5\textwidth}
      \centering
      \includegraphics[width=0.85\textwidth]{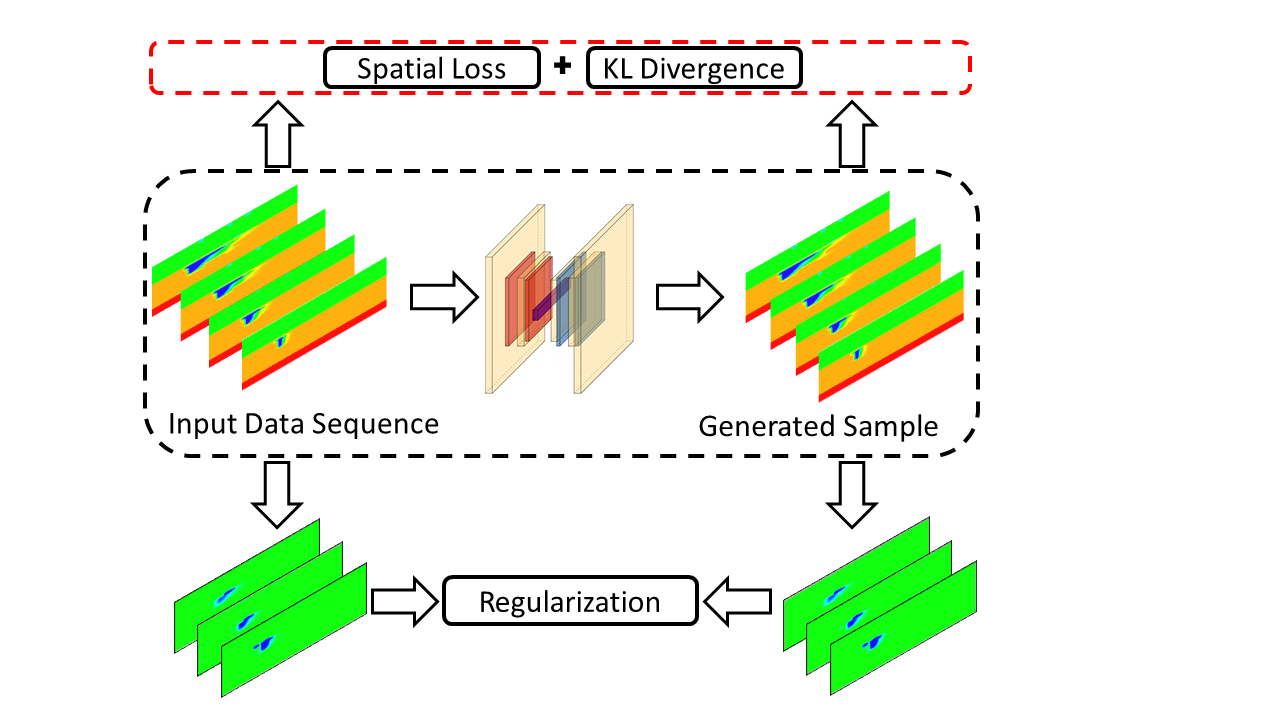} 
      \centerline{\small (b)} 
    \end{minipage}
	\caption{Schematic illustration of (a) the spatio-temporal dynamics of velocity maps at 4 consecutive times, and (b) our new variational autoencoder with  regularization. In (a), we observe CO$_2$ plume migrates towards a certain spatial direction over the time of monitoring.}
	\label{fig:VAE_Reg}
\end{figure*}
\subsection{Spatio-temporal Constrained Generative Models}

\subsubsection{Variational Autoencoder with Perception Loss}

Perception of the generated velocity map is an important criterion to evaluate the quality of the synthesized image data. For both loss functions in Eqs.~\eqref{eq:AE_Loss} and \eqref{eq:VAE}, we employ $\mathcal{L}_2$ norm to quantify the error. However, as pointed out by \cite{Zhang-2018-Unreasonable}, classic per-pixel measures would be insufficient for assessing structured data such as images. It is obvious that the perception of the generated velocity map and that of the true velocity map should be consistent throughout the whole dataset. Inspired by recent work on style transfer~\cite{gatys2015neural}, we can quantify perception using features extracted from pre-trained VGG-19 classification network~\cite{simonyan2014very}, and further calculate the perception error between the true and the generated velocity maps. Thus, reducing the perception error can make our generated velocity maps more physically realistic.  

Our new VAE generative model is shown in Fig.~\ref{fig:VAE_Style}(a), where one additional loss function (i.e., ``Perception Loss'') is added on top of the spatial loss and KL divergence. In Fig.~\ref{fig:VAE_Style}(b), we illustrate how we extract spatial features and use them to calculate the  perception loss. Particularly, we generate representation using the Gram matrix, $G^{l}\in\mathbb{R}^{N_l\times N_l}$, on feature maps from several intermediate layers of VGG-19 net
\begin{equation}
    G_{ij}^{l}=\sum_{k}F_{ik}^{l}F_{jk}^{l},
    \label{eq:GramMatrix}
\end{equation}
where $G_{ij}^{l}$ is the inner product between the vectorized feature $i,j$ at layer $l$ and $F_{ik}^{l}$ is the position $k$ of the vectorized feature map of the $i^{th}$ filter at layer $l$ of VGG-19 net. There are $N_l$ feature maps at layer $l$ of VGG-19 net with the size of $M_l$ which is the height times the width of the feature map. With the Gram matrix of ground truth velocity map, $x$~(denoted as ``$G^l$'') and that of the generated velocity map, $\hat{x}_{\alpha}$~(denoted as ``$A^l$'') obtained at layer $l$, we will have the perception loss function as 
\begin{equation}
    L_{phys}=\sum_{l}\lambda_{l}\sum_{ij}(G_{ij}^{l}-A_{ij}^{l})^{2},
    \label{eq:physicsloss}
\end{equation}
where $\lambda_{l}=\frac{1}{4N_{l}^{2}M_{l}^{2}}$ is the coefficient of the  perception loss at layer $l$. Hence, our new VAE with  perception loss function becomes
\begin{equation}
\begin{split}
    \mathcal{L}(\theta,\phi) &= \mathcal{L}_{recon} + \mathcal{L}_{kld} + \mathcal{L}_{phys},\\
    &=\sum_{i}(x_{i}-\hat{x}_{i})^{2}+D_{KL}(q_{\theta}(z_{i}|x_{i})\|p(z_{i}))\\&+\sum_{l}\lambda_{l}\sum_{ij}(G_{ij}^{l}-A_{ij}^{l})^{2}.
\end{split}
    \label{eq:VAE_physicsloss}
\end{equation}
An important hyper-parameter that needs to be carefully tuned is the selection of layers from VGG-19 net that will be used for calculating the  perception loss. We will provide more details later in the numerical test.

\begin{figure*}[ht!]
    \centering
    \includegraphics[width=0.95\textwidth]{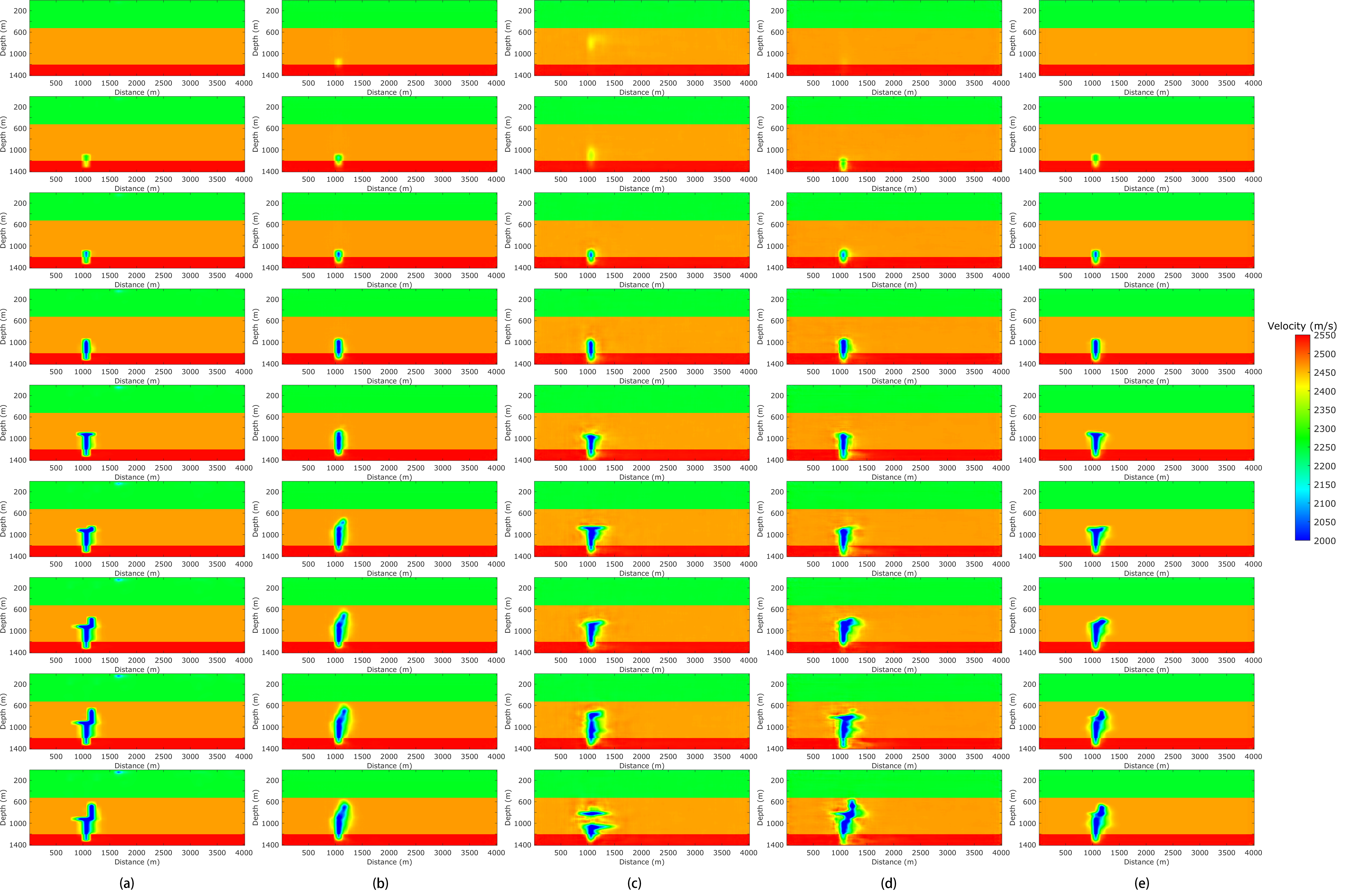}
    \caption{(a)~Ground truth velocity maps, and generated velocity maps using (b)~autoencoder, (C) VAE, (d) VAE with  perception loss, and (e) VAE with regularization.}
    \label{fig:ori gen comp}
\end{figure*}

\subsubsection{Variational Autoencoder with Regularization} \label{vae_reg}

As we discussed previously, prominent phenomena also play a critical role in designing our generative model. For the CO$_2$ storage problem, it has been well understood that starting from the injection well, CO$_2$ will enter the formation at high flow rates and migrate relatively and vigorously into the most permeable regions under strong pressure gradients while displacing native fluids~(e.g., brine)~\cite{Birkholzer-2015-co2}. Our full-physics simulations as shown in Fig.~\ref{fig:kimberlina} accurately illustrate this process, which results in a prominent spatially and temporally varying pattern of the CO$_2$ plume~(shown in Fig.~\ref{fig:VAE_Reg}(a)). We expect our generative model would respect~(at least not violate) this particular dynamics. To enforce this constraint, our idea is to design a regularization term informed by the leakage process with the hope to produce new samples being consistent with the underline spatio-temporal dynamics. 

As shown in Fig.~\ref{fig:VAE_Reg}(a), we observe that the CO$_2$ plume would span towards a certain spatial direction over the time during the migration, which indicates a clear spatio-temporal dynamical pattern. We, therefore, impose regularization on top of the difference between velocity maps at two consecutive times and ensure the dynamics of the ground truth would be preserved to its best when generating new synthetics. To achieve this, we employ a $\mathcal{L}_1$-norm based regularization given by
\begin{equation}
    \mathcal{L}_{reg}= \left \|(x_{s,t1}-x_{s,t2}) - (\hat{x}_{s,t1}-\hat{x}_{s,t2}) \right \|_{1},
    \label{eq:regularization}
\end{equation}
where $x_{t1}$ and $x_{t2}$ are ground truth velocity maps at two consecutive times, respectively. Accordingly, $\hat{x}_{t1}$ and $\hat{x}_{t2}$ are generated velocity maps at the same times, respectively. The times, $t1$ and $t2$, are adjacent to each other with $t1>t2$. The reason that we use $\mathcal{L}_1$ norm instead of $\mathcal{L}_2$ norm is that the value of the subtraction of the differences of two consecutive velocity maps can sometimes be very close to zero, which would make the $\mathcal{L}_2$-norm value too small and may lead to the gradient vanishing issue. In Fig.~\ref{fig:VAE_Reg}(b), we provide the network structure of our generative model using a new VAE loss function with regularization
\begin{equation}
\begin{split}
    \mathcal{L}(\theta,\phi) &= \mathcal{L}_{recon}+\mathcal{L}_{kld}+ \gamma\, \mathcal{L}_{reg},\\
    &=\sum_{i}(x_{i}-\hat{x}_{i})^{2}+D_{KL}(q_{\theta}(z_{i}|x_{i})\|p_{\phi}(z_{i}))\\&+\gamma\,\left \|(x_{s,t1}-x_{s,t2}) - (\hat{x}_{s,t1}-\hat{x}_{s,t2})\right \|_{1},
\end{split}
    \label{eq:VAE_reg}
\end{equation}
where the first and the second terms are the reconstruction loss and KL-divergence of the VAE, respectively. The third term is regularization. $\gamma$ is the regularization parameter. The regularization parameter in Eq.~\eqref{eq:VAE_reg} is important to the accuracy of data generation. We will explore its impact and how we select it in our numerical test. Another technical details may be worthwhile mentioning is the selection of the norms~(i.e., $\mathcal{L}_2$ norm versus $\mathcal{L}_1$ norm) in our loss functions and the regularization terms. The $\mathcal{L}_2$ loss~(i.e., Mean Squared Error~(MSE)) and $\mathcal{L}_1$ loss~(i.e., Mean Absolute Error~(MAE)) are two of the most popularly used functions. Since MAE is minimized by conditional median which may lead to bias during optimization, we therefore choose MSE, which is minimized by conditional mean. However, we select the $\mathcal{L}_1$ regularization term to promote the sparsity of the coefficients when constraining the differences between two (or thee) adjacent velocity maps. We provide numerical test to compare the performance using different loss functions in the Supplement. 

\begin{figure}
    \centering
    \includegraphics[width=1\linewidth]{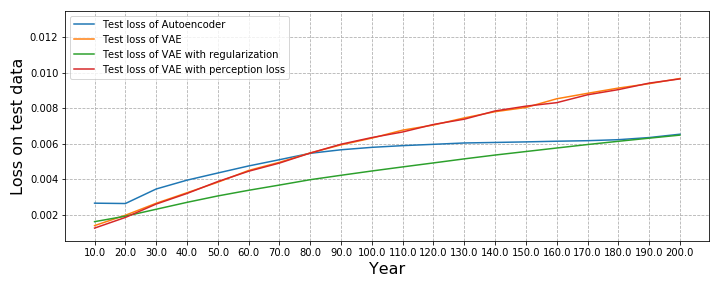}
    \caption{Reconstruction loss on test dataset using different models w.r.t. each year in the dataset.}
    \label{fig:testlosscomp}
\end{figure}

\begin{figure*}[h]
    \centerline{
    \subfloat[]{\includegraphics[width=.24\linewidth]{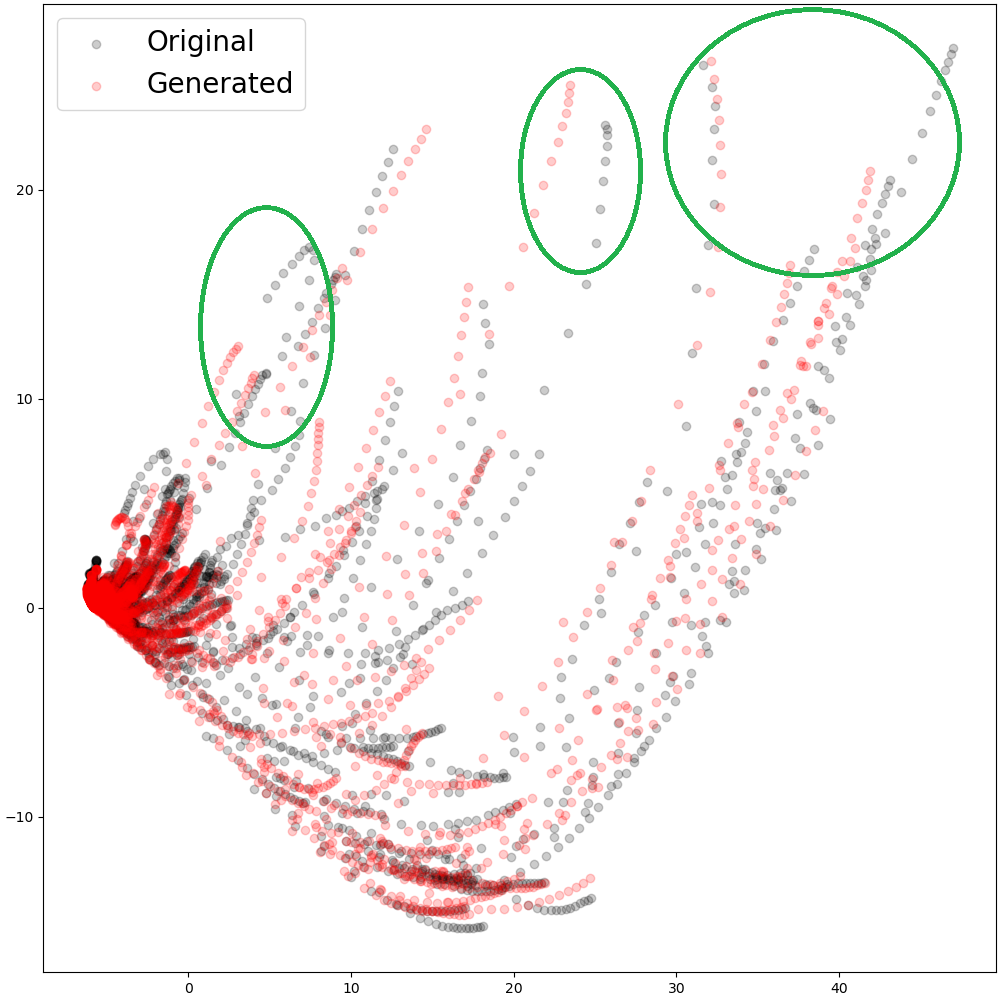}}
    \quad
    \subfloat[]{\includegraphics[width=.24\linewidth]{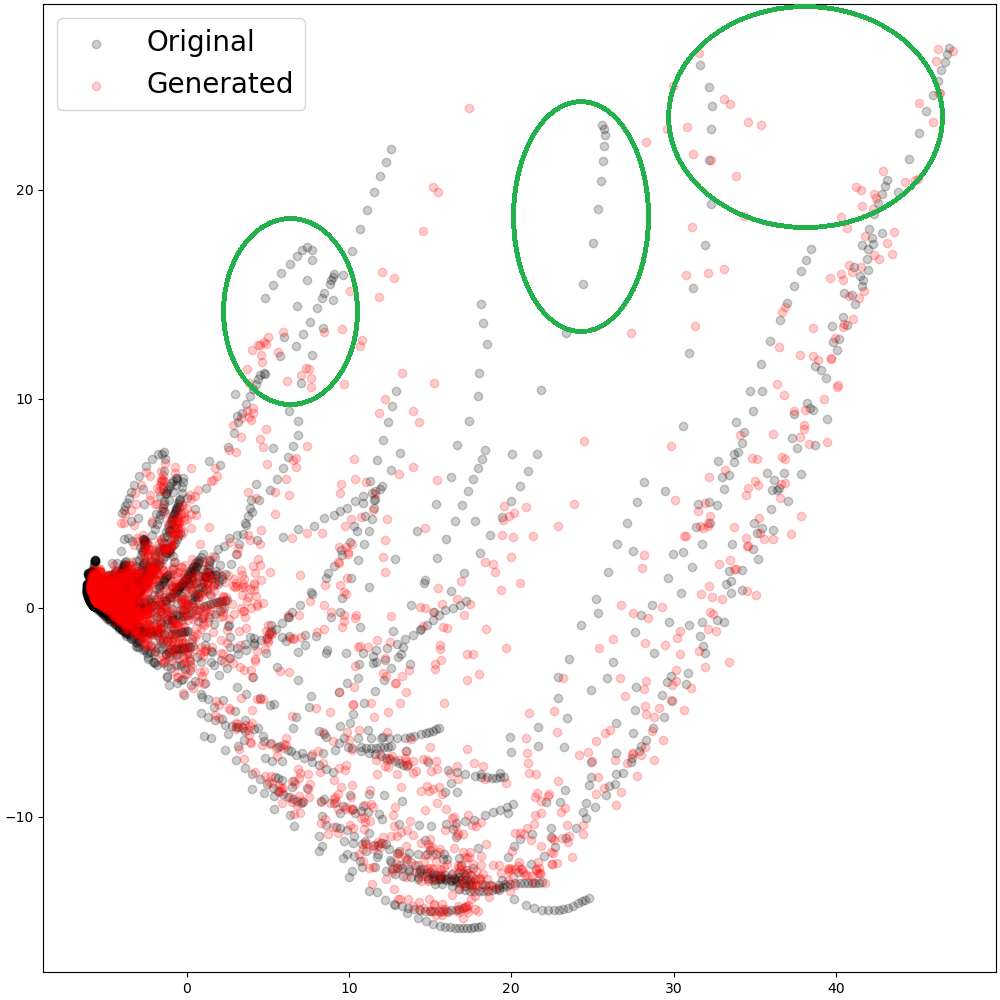}}
    \quad
	\subfloat[]{\includegraphics[width=.24\linewidth]{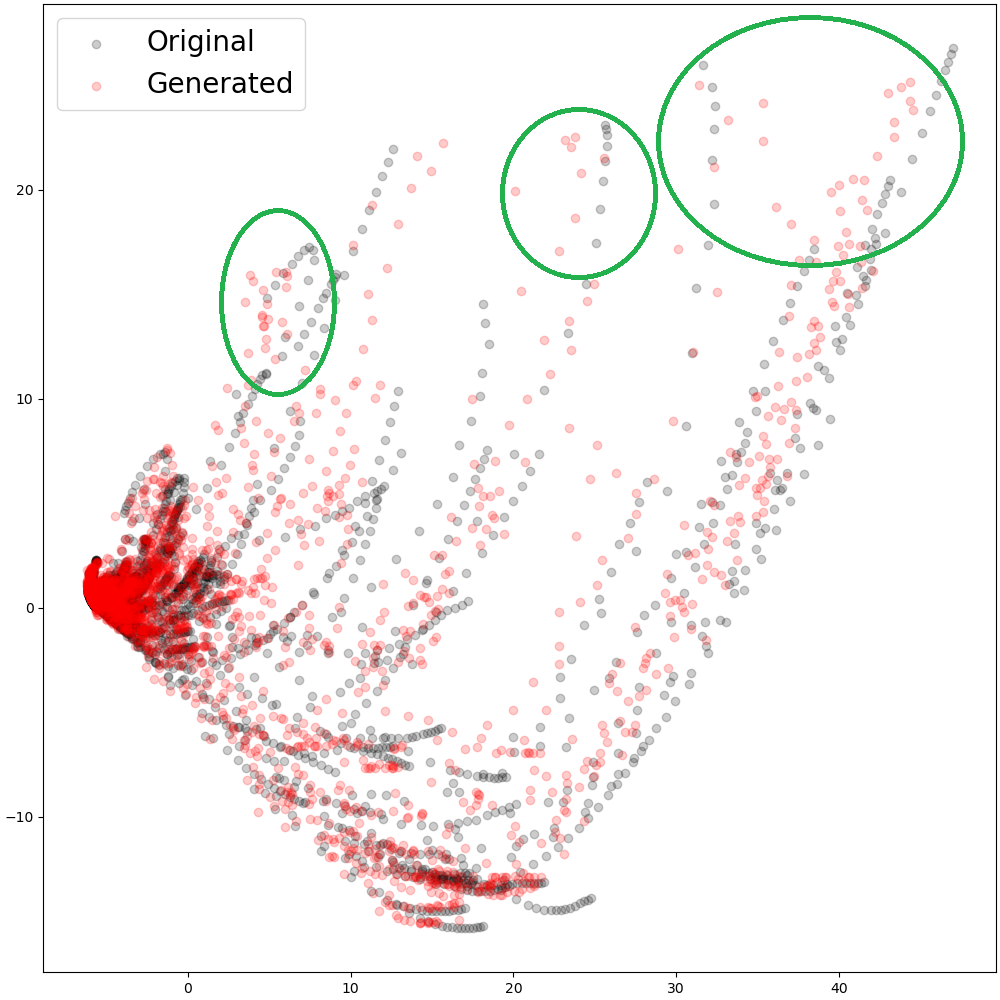}}
	\quad
	\subfloat[]{\includegraphics[width=.24\linewidth]{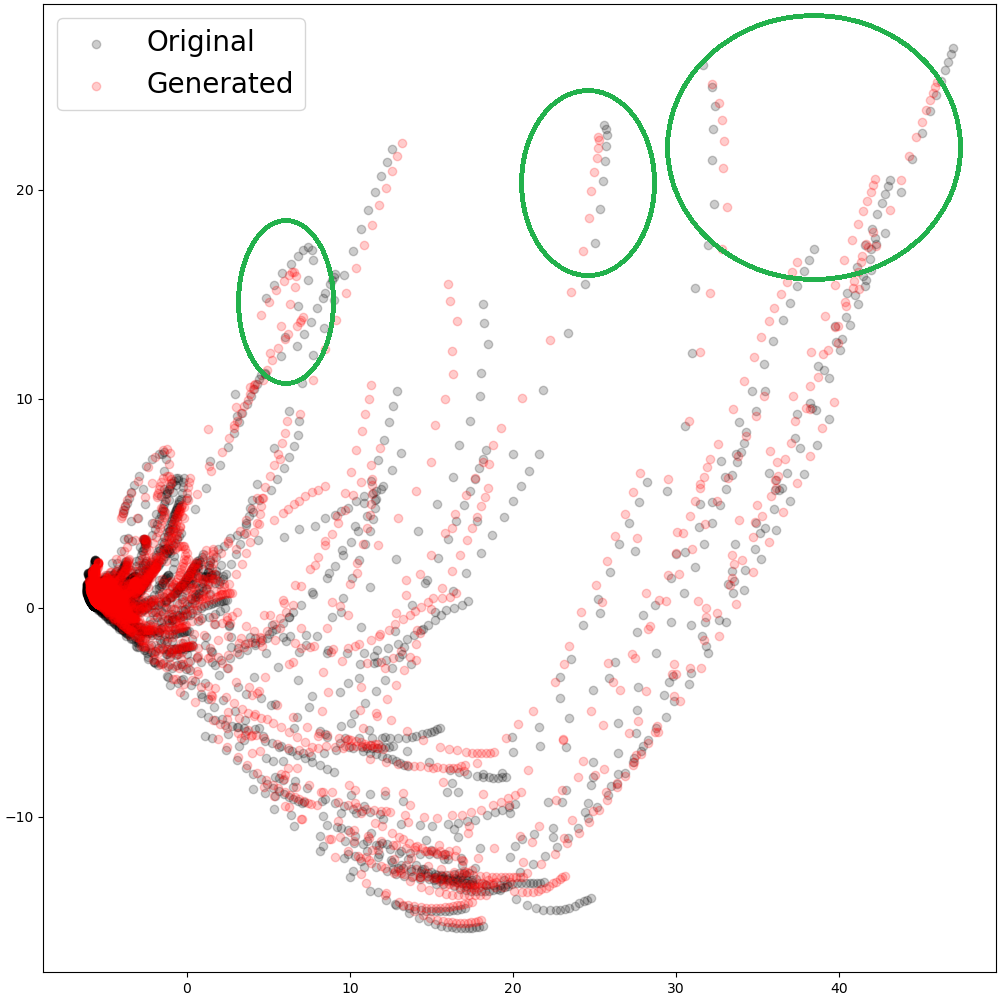}}}
	    \centerline{
	\subfloat[]{\includegraphics[width=.24\linewidth]{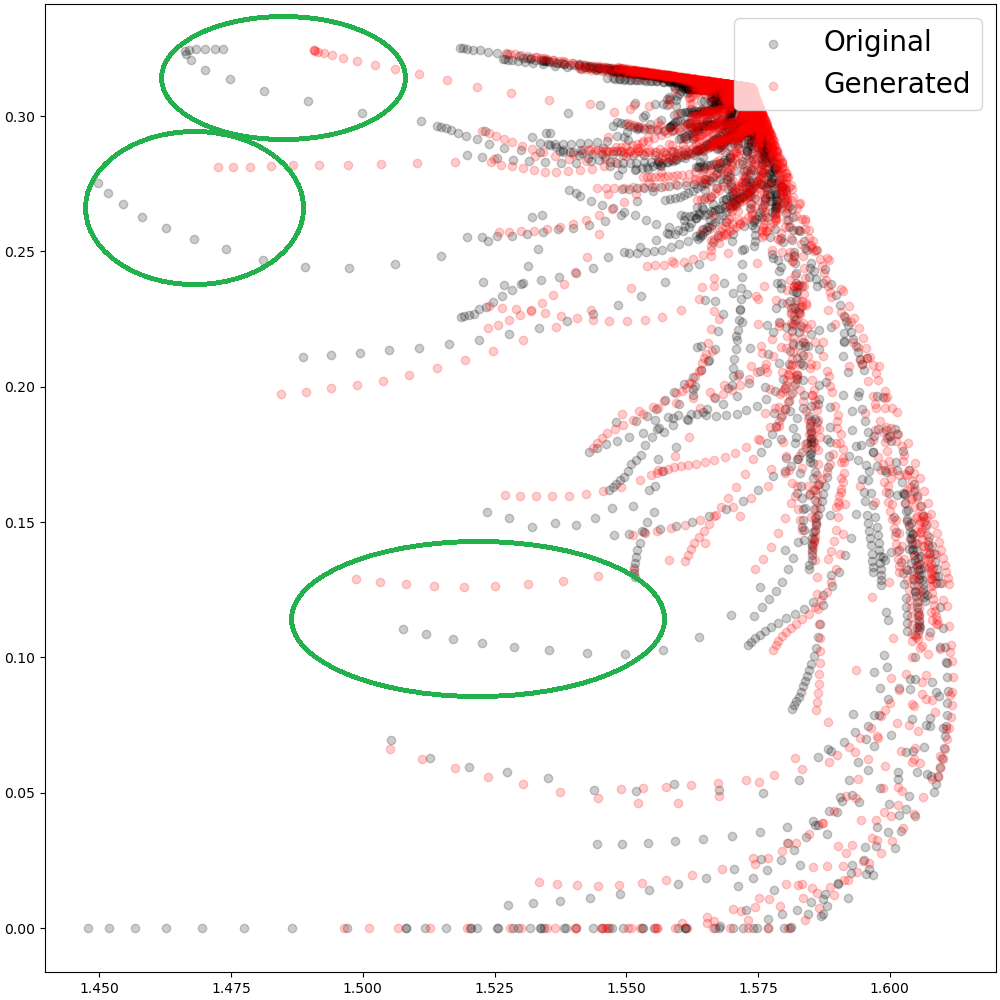}}
	\quad
	\subfloat[]{\includegraphics[width=.24\linewidth]{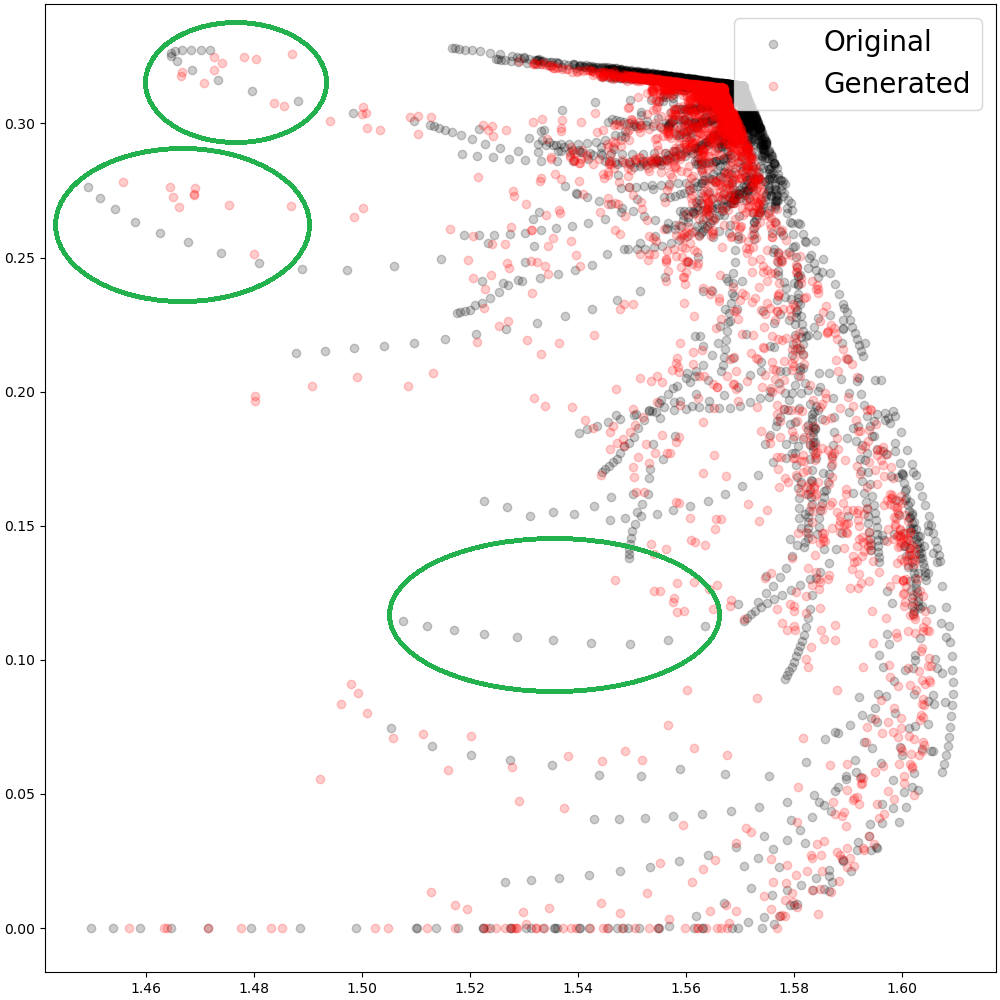}}
	\quad
    \subfloat[]{\includegraphics[width=.24\linewidth]{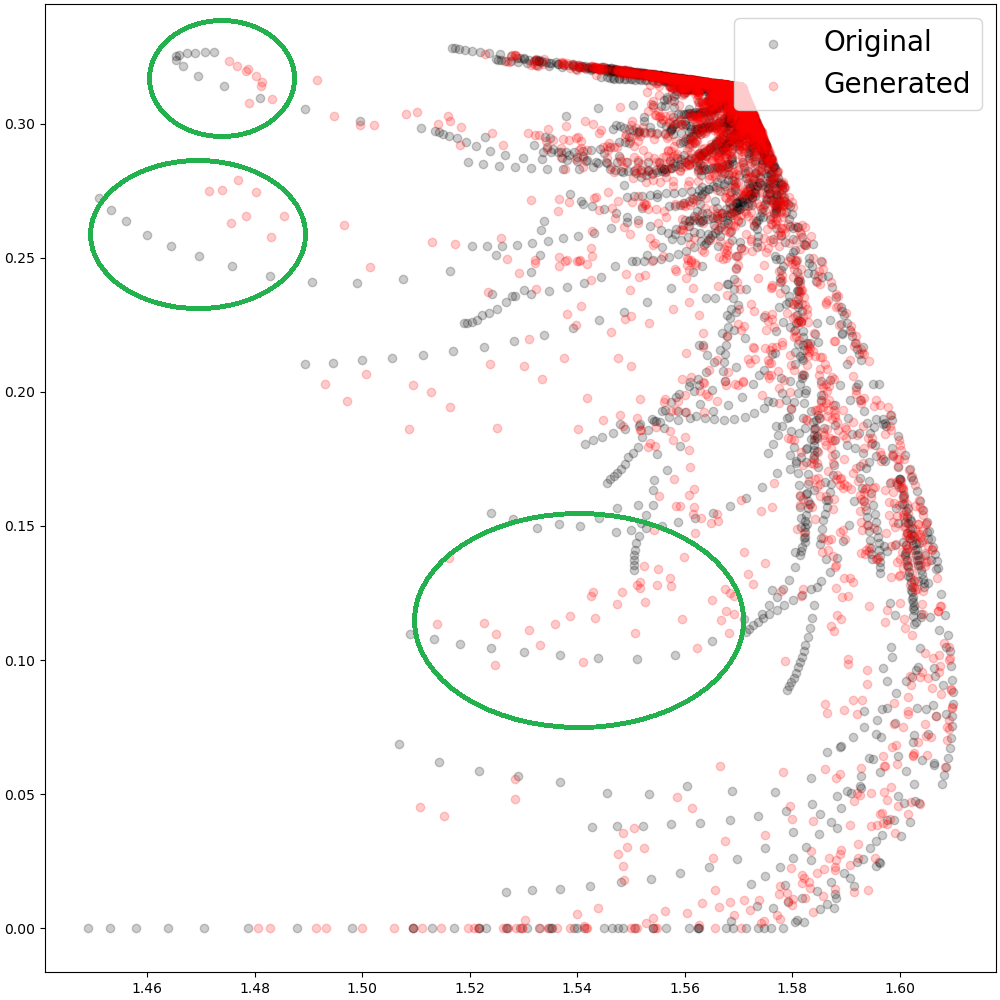}}
    \quad
    \subfloat[]{\includegraphics[width=.24\linewidth]{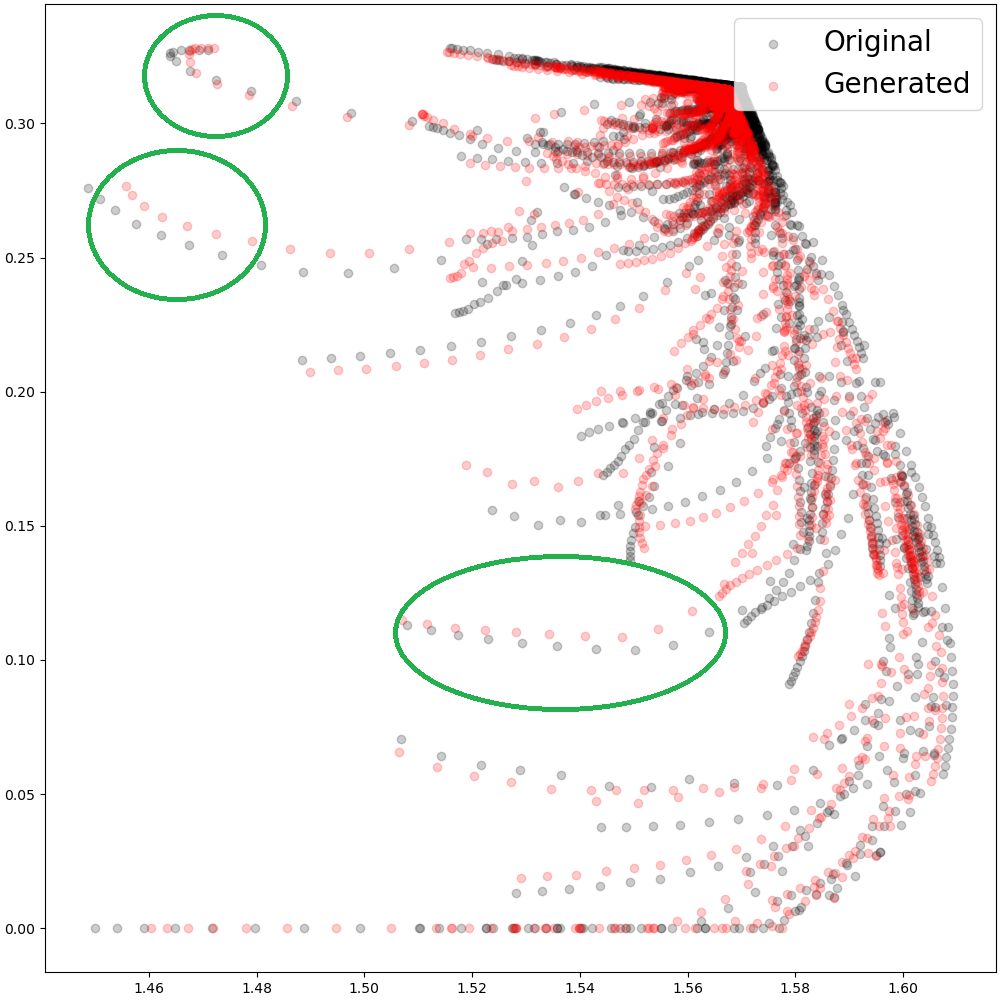}}
    }
    \caption{Clustering results of the generated versus true samples. PCA-based clustering for (a) autoencoder, (b) VAE, (c) VAE with perception loss, and (d) VAE with regularization; NMF-based clustering for (e) autoencoder, (f) VAE, (g) VAE with perception loss, and (h) VAE with regularization.}
    \label{fig:clustering_results}
\end{figure*}

\section{Experiments}
\label{sec:results}

With all 4 different generative models being designed in the previous section, we will validate their performances in a couple of scenarios including a general assessment of the synthesized data~(Test 1) and the performance in imaging small CO$_2$ leakage~(Test 2). We will also provide numerical tests to illustrate what would be a reasonable augmented data size~(Test 3) and how we pick some of the critical hyper-parameters~(Test 4).    

\subsection{Experiment Setup} 
\label{exp setup}

We use 800 leakage scenarios in our dataset~(16,000 samples in total) as a training dataset and the rest of the simulations~(3,763 samples in total) as a test dataset. We provide in the Supplement the details of the generation of the synthetic seismic data. For training of our proposed data augmentation models, we use a batch size of 32, and train models for 100 epochs using ADAM optimizer with a learning rate of 0.0001. The initialization of model weights is based on He initialization~\cite{he2015delving}. We provide the training curves of our different generative models in the Supplement. For the training of InversionNet, we use a batch size of 24, and train InversionNet for 80 epochs using ADAM optimizer with an initial learning rate of 0.01, and with a weight decay coefficient of 0.0001. We also provide the network structure of InversionNet for readers who may be interested in the Supplement.

    \begin{table*}[t]
    \centering
    \begin{tabular}{c|c|c|c|c}
        \hline
                       & Autoencoder & VAE & VAE\_percep & VAE\_reg \\
        \hline
        Parameter \# & 3,382,290 & 6,432,818 & 6,432,818 & 6,432,818 \\
        \hline
        GPU Memory Cost & 10.3GB & 14.6GB & 14.6GB & 14.6GB  \\
        \hline
        Time (Training/epoch) & 48s & 36s & 258s & 48s \\
        \hline
        Time (Training/total) & 80m & 60m & 430m & 80m \\
        \hline
        Time (Generation/sample) & 1.36s & 4.38s & 4.38s & 4.38s \\
        \hline
        Time (Generation/set) & 3.13m & 3.41m & 3.41m & 3.41m \\
        \hline
    \end{tabular}
    \caption{Computational costs of different generative models. Row~1 is the size of the each model. Row~2 is the memory cost. Row~3 is the time per epoch in training each model. Row~4 is the total time in training each model. Row~5 is the time in generating a single sample. Row~6 is time in generating 3,000 velocity maps set in parallel.}
    \label{tab:comp_cost}
    \end{table*}

\begin{table*}[h]
    \centering
    \begin{tabular}{c|c|c}
        \hline
         & InversionNet without augment & InversionNet with augment \\
         \hline
        Training Time (total) & 1h23m & 1h31m \\
        \hline
    \end{tabular}
    \caption{Computational costs of training InversionNet without and with augmented dataset with 3,000 more velocity maps.}
    \label{tab:invnet_time}
\end{table*}
\subsection{Test 1: Velocity Map Generation}\label{test1}

We provide the synthesized velocity maps generated using our four different generative models in Fig.~\ref{fig:ori gen comp}. For VAE, we use the first column, the ground truth velocity maps as the input velocity maps. For autoencoder, we use two velocity maps of 10-year and 200-year as the input velocity maps. Overall, we observe that all four generative models produce reasonable results. VAE (Fig.~\ref{fig:ori gen comp}(c)) yields images with the highest variability among all four models. This is particularly true when comparing to autoencoder results~(Fig.~\ref{fig:ori gen comp}(b)). However, due to the variability, some unrealistic features can be also observed in the VAE results. An example would be the one at the last row, where the whole leakage plume is unphysical split into two. The VAE with perception loss~(Fig.~\ref{fig:ori gen comp}(d)) produces better results in preserving the perception of the velocity map. The unphysical data in the later stage is improved. Meanwhile, the images at the early stage also match better to the true images comparing to the VAE results. However, we also notice some artifacts being generated in the VAE with the perception loss model. The best results produced by all four models is the VAE with regularization~(Fig.~\ref{fig:ori gen comp}(e)). It produces not only cleaner images but also highly accurate images in the early stage. To further quantitatively compare different generative models, we run our models on the test dataset and calculate the test loss w.r.t. each year of the data. The result is shown in Fig.~\ref{fig:testlosscomp}. Consistent with what we observe in Fig.~\ref{fig:ori gen comp}, VAE with regularization~(in red), yields the best performance among all four models. Autoencoder yields a lower reconstruction loss for velocity maps after 80 years compared to those of vanilla VAE, VAE\_{percep}, and VAE\_{reg} models. However, this only implies that autoencoder produces a better performance on reconstructing velocity maps after 80 years, and it does not mean that autoencoder can generate more realistic and diversified velocity maps from the underlying distribution, which however is essential for InversionNet to capture the underlying data distribution and help with the generalization ability. We also provide the difference of the synthesized velocity maps in Fig.~\ref{fig:ori gen comp} to their corresponding ground truth velocity maps in the Supplement.

Another means to justify the quality of our synthesized data is to visualize the distribution of the generated data versus that of the true data. To achieve this, we employ two commonly used methods: Principal Components Analysis~(PCA)~\cite{smith2002tutorial} and Non-Negative Matrix Factorization (NMF)~\cite{lee1999learning}. The visualization results are provided in Fig.~\ref{fig:clustering_results}. Regardless of the clustering approaches, the VAE with regularization~(Figs.~\ref{fig:clustering_results}(d) and (h)) produces the distribution matching most closely to that of the ground truth. VAE and VAE with perception loss models yield comparable results. The autoencoder-based model performs the worst out of all four models.

Computation cost is also an important factor in evaluating the performance of a generative model. To that perspective, we provide in Table~\ref{tab:comp_cost} more details of the cost by comparing the model complexity (number of parameters), memory consumption, training time per epoch, total training time, sample-generation time, and total generation time required. Particularly, we compare all four models listed in our manuscript including autoencoder, VAE, VAE\_{percep} and VAE\_{reg}. We observe that autoencoder yields the smallest number of network parameters among all four models and all the other three VAE-based models yield comparable network complexity and memory requirement. As for the training time cost, VAE\_{percep} is the most time-consuming whereas the remaining three models are comparable in training cost. The excessive time of training the VAE\_percep model is due to accessing multiple layers of VGG-19 for extracting relevant spatial features that would be needed in computing the Gram matrix and perception loss. Once the models are fully trained, generating the augmented dataset~(3,000 samples) only spends around 4 minutes, which is not very time-consuming. With new samples being generated, we provide the training time comparison of InversionNet with/without augmented dataset in Table~\ref{tab:invnet_time}. We notice that training InversionNet with augmentation dataset will only increase 8 minutes in training.

To summarize, in this test we demonstrate the capability of our four generative models in synthesizing high-quality velocity maps, which would provide additional data to train data-driven seismic imaging methods. Particularly, our VAE with regularization model generates the synthesized data with the most appealing visual quality and matches best to the true data distribution.

\subsection{Test 2: Performance on Edge Cases}

The main purpose of this test is to evaluate the performance of our developed data augmentation techniques in improving the data-driven seismic imaging method in characterizing small leakage. We firstly generate 4 groups of synthesized data using our proposed generative models. There are 3,000 velocity maps in each group. As a baseline, we will train InversionNet~\cite{wu-2019-inversionnet} with an initial training dataset, which consists of 800 simulations, amounting to a total of around 15,000 samples. We further design two different test categories of one with all sizes of leakage~(named as ``General Leakage'') and the other one with only tiny and small leakage~(named as ``Small Leakage''). For the definitions of different leakage~(tiny, small, medium, and large), please refer to Eq.~\eqref{eq:leakCases} and Fig.~\ref{fig:mass dis}. The test samples of different leakage sample in General Leakage and Small Leakage are provided in Table~\ref{tab:TestData}.  
\begin{table}[h]
    \centering
    \begin{tabular}{c|c|c|c|c}
        \hline
                       & Tiny & Small & Medium & Large \\
        \hline
        General leakage & 717 & 770 & 675 & 1494  \\
        \hline
        Small leakage & 153 & 38 & 0 & 0 \\
        \hline
    \end{tabular}
    \caption{Two different test sets for evaluating the performance of our generative models. For the definitions of different leakage~(tiny, small, medium, and large), please refer to Eq.~\eqref{eq:leakCases} and Fig.~\ref{fig:mass dis}.}
    \label{tab:TestData}
\end{table}

\begin{table*}[t]
    \centering
    \begin{tabular}{c|c|c|c|c|c}
        \hline
         & Baseline & AE & VAE & VAE\_percep & VAE\_reg \\
        \hline
        General leakage & 0.001294 & 0.001229 & 0.001522 & 0.001331 & \textbf{0.001093} \\
        \hline
        Small leakage & 0.000780 & 0.000813 & 0.001157 & 0.000924 & \textbf{0.000646} \\
        \hline
    \end{tabular}
    \caption{Test loss of InversionNet on General Leakage and Small Leakage tests without augmentation~(Col~2), and with augmentation data generated using autoencoder~(Col~3), VAE~(Col~4), VAE with perception loss~(Col~5) and VAE with regularization~(Col~6). The results using VAE with regularization are the best comparing to all others.}
    \label{tab:invtest}
\end{table*}

\begin{figure}[ht]
\centerline{
    \includegraphics[width=0.6\linewidth]{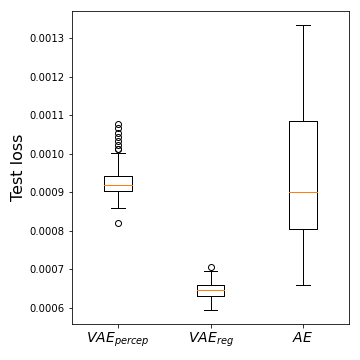}}
\caption{Illustration of the error distribution~(median, range, and outliers) in a box-plot on the small leakage test.}
\label{fig:boxplot_smallleak}
\end{figure}

For all tests, we train InversionNet for 80 epochs to assure its convergence. We report in Table~\ref{tab:invtest} the test loss of InversionNet on both testing categories with/without augmented training data sets. Particularly, our VAE with regularization yields the smallest loss value for both ``General Leakage'' and ``Small Leakage''. On the other hand, VAE model~(Col~4) produces the worst results among all four methods. We suspect that high variability and some unphysical synthesized samples ``confuses'' the InversionNet in learning the data distribution leading to degraded performance. This can be confirmed by noticing that with some additional constraints being imposed to the VAE model, an immediate performance improvement can be observed in the results using VAE with perception loss~(Col~5) and VAE with regularization~(Col~6). To better understand the error distribution using different generative models, we also provide in Fig.~\ref{fig:boxplot_smallleak} a box-plot on the small leakage test. Out of the three methods, it is clear that VAE\_reg yields the best performance with the smallest median value and interquartile ranges. Comparing VAE\_percep and AE models, although they both produce similar median values, the VAE\_percep is much less dispersed than the AE model. However, we notice more outliers are existing in the VAE\_percep box-plot than those in the other two box-plots, which explains degradation of the overall test loss of VAE\_percep as reported in Table~\ref{tab:invtest}. 

\begin{figure*}[h!]
    \centering
    \includegraphics[width=0.95\textwidth]{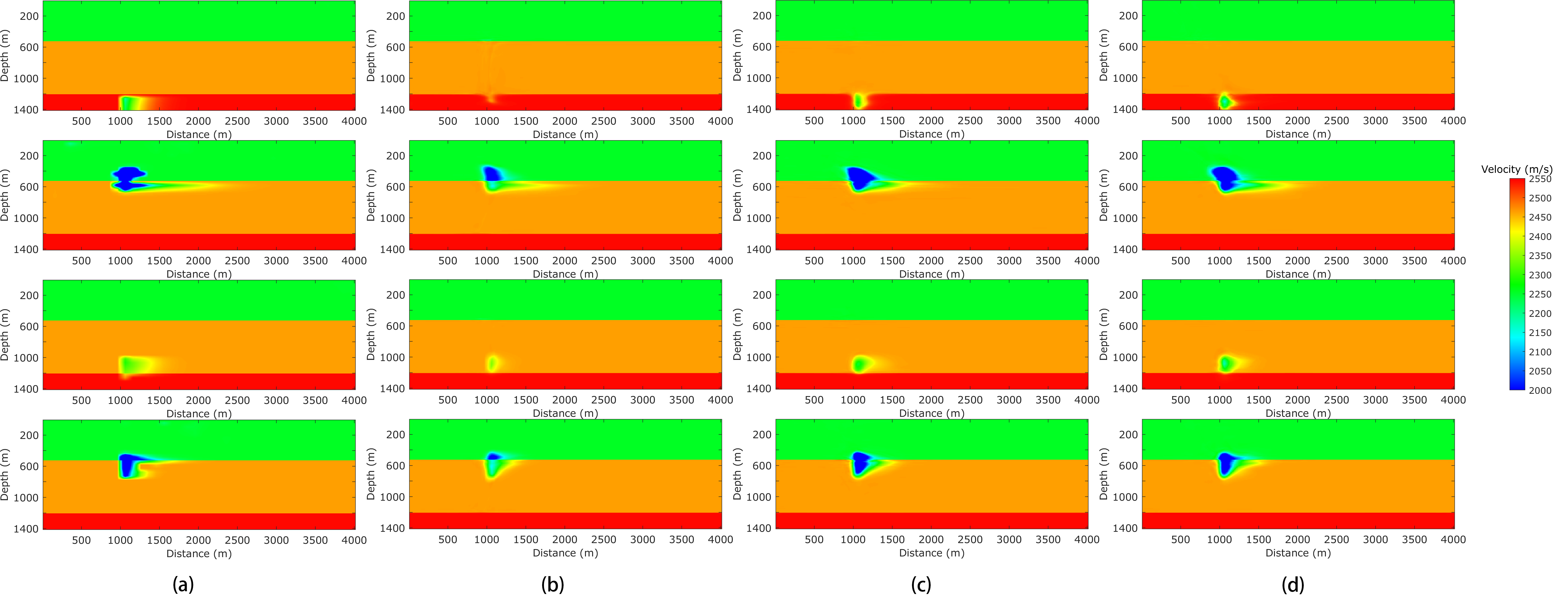}
    \caption{Four groups of InversionNet imaging results~(b, c, d) on small leakage test data. (a)~Ground truth, InversionNet imaging results (b) without augmentation, with augmented data set generated using (c) VAE with perception loss, and (d) VAE with regularization.}
    \label{fig:smalltest}
\end{figure*}

\begin{figure*}[h!]
    \centering
    \includegraphics[width=0.8\textwidth]{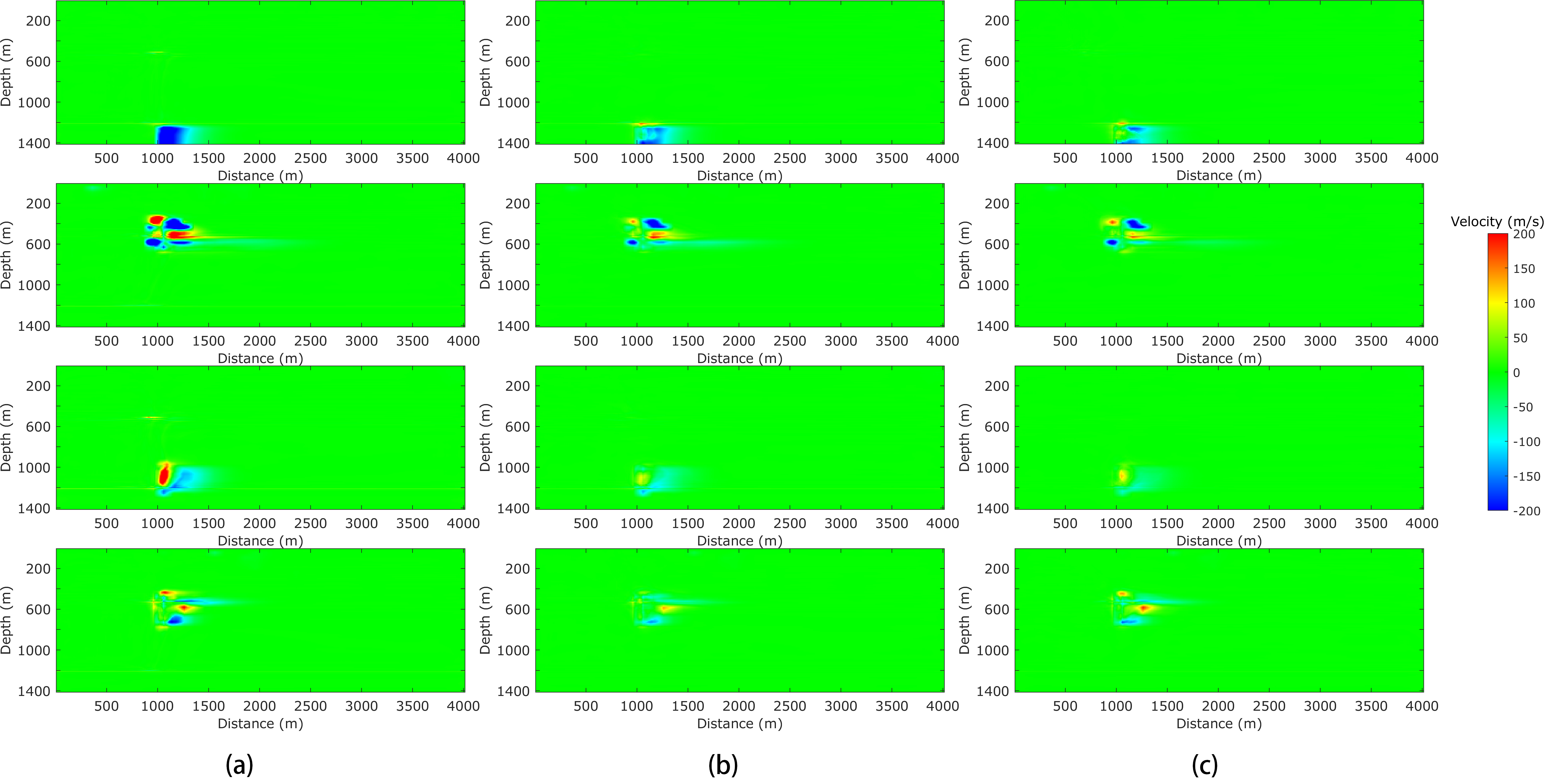}
    \caption{Four groups of differences of InversionNet imaging results to ground truth~(e, f, g) on small leakage test data. (a)~Difference of InversionNet imaging results without augmentation (Fig.~\ref{fig:smalltest} (b)) to ground truth (Fig.~\ref{fig:smalltest} (a)), (b) difference of results with augmented data set generated using VAE with perception loss (Fig.~\ref{fig:smalltest} (c)) to ground truth, (c) difference of results with augmented data set generated using VAE with regularization (Fig.~\ref{fig:smalltest} (d)) to ground truth. }
    \label{fig:test_diff}
\end{figure*}

To better visualize the performance in imaging small leakage, we provide the reconstructed imaging results of InversionNet in Figs.~\ref{fig:smalltest}(b) to (d). The differences of the reconstructions~(by subtraction results from ground truth) are further provided in Figs.~\ref{fig:test_diff}(a) to (c). The ground truth of the testing samples is shown in Fig.~\ref{fig:smalltest}(a). To quantify the errors of the imaging results, we use two metrics, the mean-absolute errors (MAE) and structural similarity indexes~(SSIM)~\cite{wang2004image}. The leakage mass and errors of the imaging results are provided in Table~\ref{tab:mae ssim}. InversionNet trained without any augmentation data yields the worst imaging results. The CO$_2$ plume is either very hard to visualize or distorted severely, which results in the highest MAE and the lowest SSIM value comparing to others. InversionNet trained on augmented data sets produce much-improved imaging results. Particularly, the one using VAE with  regularization yields the best imaging results with the smallest MAE and highest SSIM values.

\begin{table*}[t]
    \centering
    \begin{tabular}{c|c|c|c|c|c}
        \hline
        Group & Leakage Mass & Metric & Baseline & VAE\_percep & VAE\_reg \\
        \hline
        \multirow{2}{*}{1} & \multirow{2}{*}{$4.08\times10^{6}~\mathrm{kg}$} & MAE & 0.00277 & 0.00237 & \textbf{0.000889} \\
        \cline{3-6}
         &  & SSIM & 0.9907 & 0.9929 & \textbf{0.9936} \\
        \hline
        \multirow{2}{*}{2} & \multirow{2}{*}{$4.29\times10^{3}~\mathrm{kg}$} & MAE & 0.00437 & 0.00346 & \textbf{0.00290} \\
        \cline{3-6}
         & & SSIM & 0.9833 & 0.9842 & \textbf{0.9871} \\
         \hline
        \multirow{2}{*}{3} & \multirow{2}{*}{$8.89\times10^{5}~\mathrm{kg}$} & MAE & 0.00102 & \textbf{0.000742} & 0.000766 \\
        \cline{3-6}
         &  & SSIM & 0.9985 & \textbf{0.9987} & 0.9986 \\
        \hline
        \multirow{2}{*}{4} & \multirow{2}{*}{$8.05\times10^{5}~\mathrm{kg}$} & MAE & 0.00255 & \textbf{0.00138} & 0.00168 \\
        \cline{3-6}
         &  & SSIM & 0.9926 & 0.9928 & \textbf{0.9949} \\
        \hline
        
    \end{tabular}
    \caption{Leakage mass~(Col~2) of CO$_2$ of four groups of velocity maps showed in Fig.~\ref{fig:smalltest}. MAE and SSIM errors of InversionNet imaging results using baseline without augmentation~(Col~4), VAE with perception loss~(Col~5) and VAE with regularization~(Col~6).}
    \label{tab:mae ssim}
\end{table*}

\begin{figure*}[h]
\centerline{
\subfloat[]{%
    \includegraphics[width=0.23\linewidth]{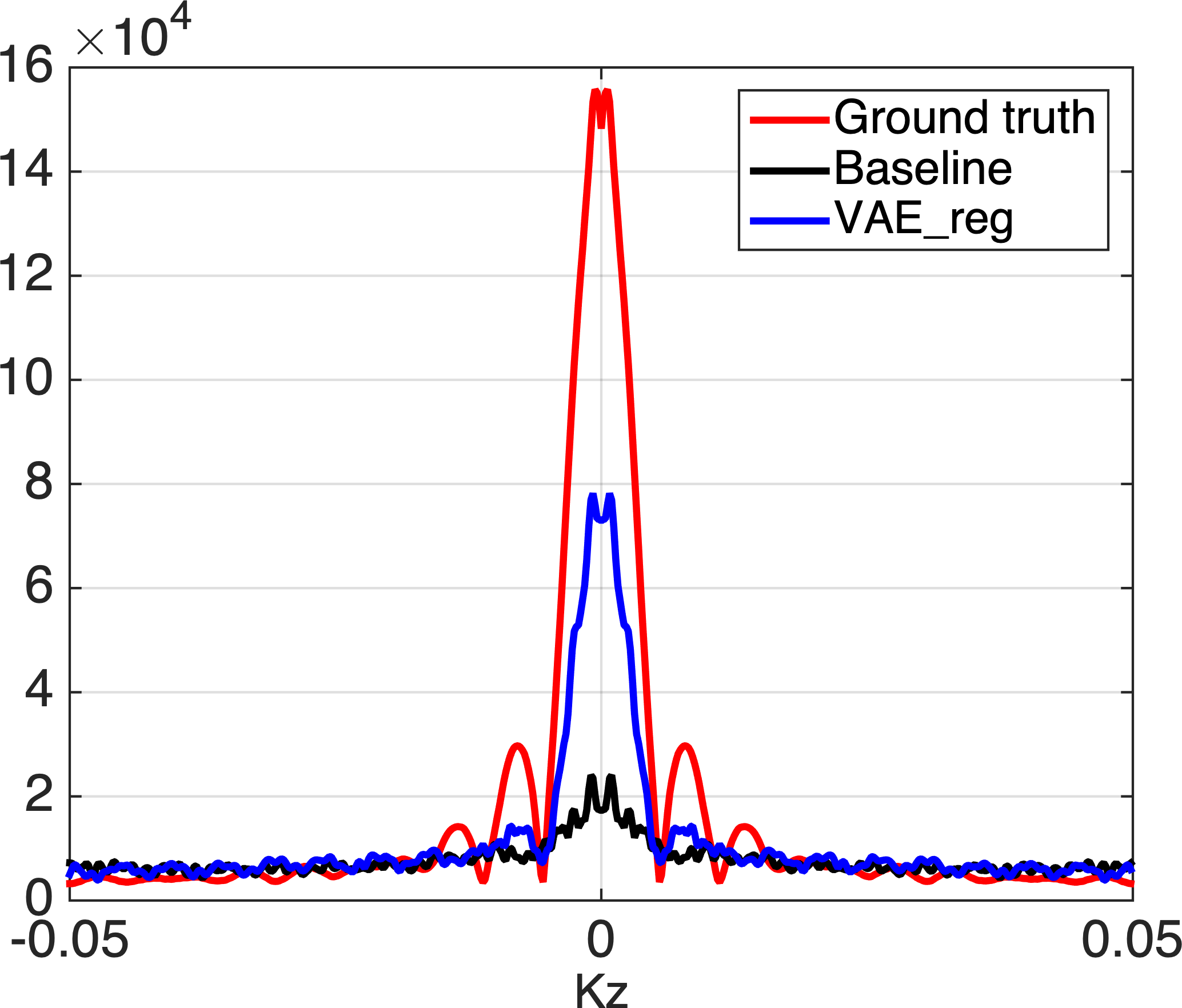}%
    \label{fig:row1}}
    \quad 
\subfloat[]{%
    \includegraphics[width=0.23\linewidth]{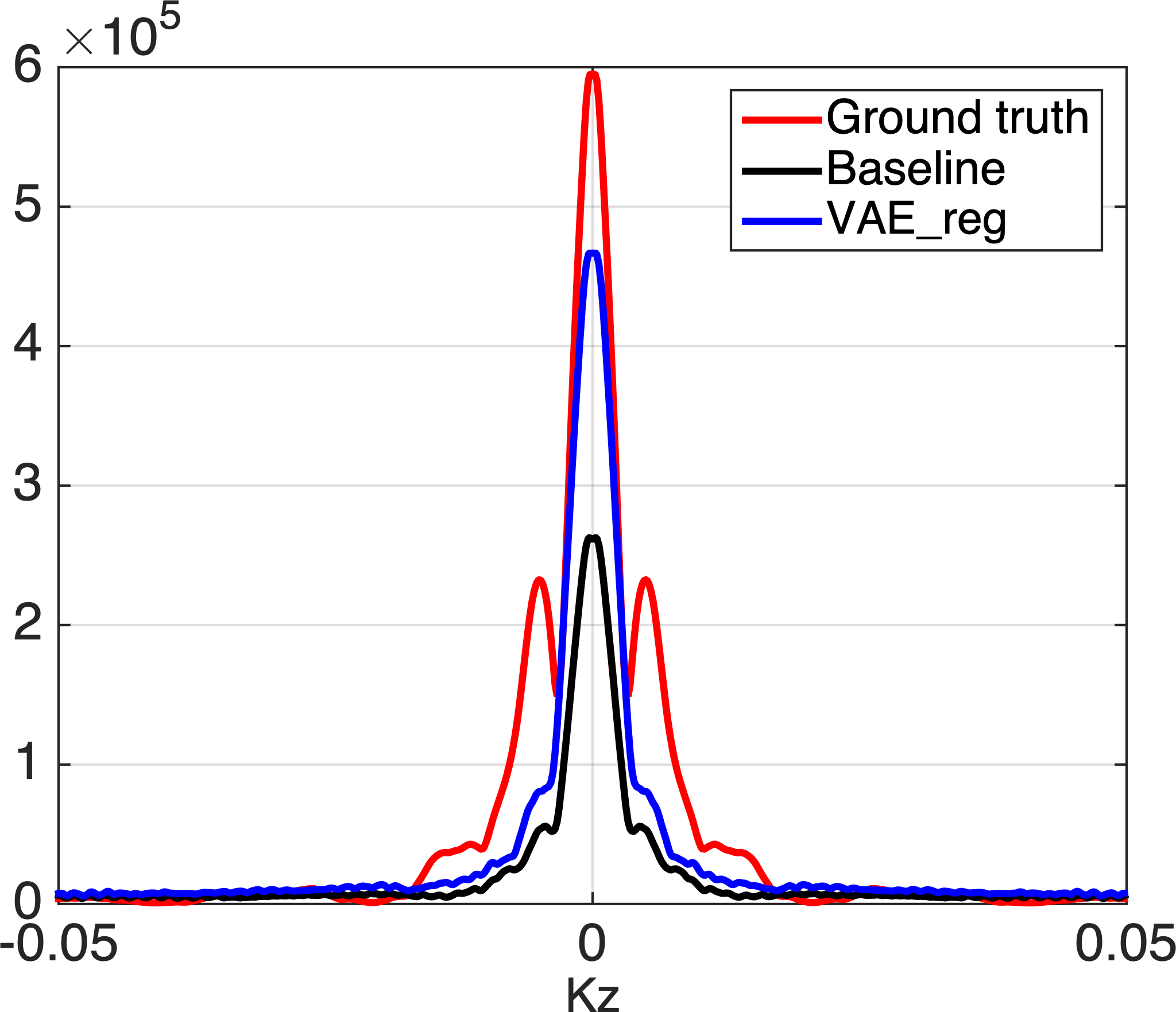}%
    \label{fig:row2}}
    \quad 
\subfloat[]{%
    \includegraphics[width=0.23\linewidth]{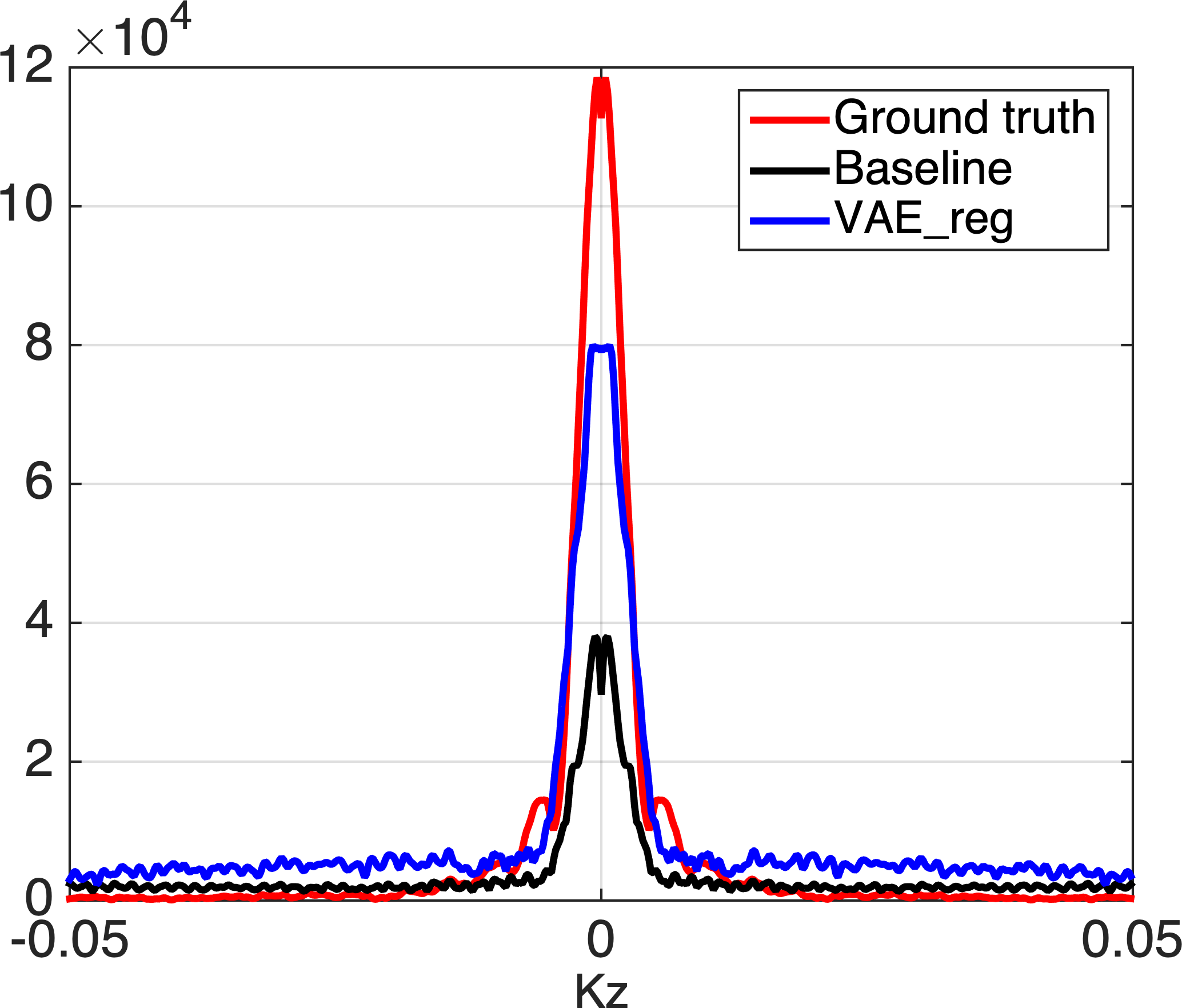}%
    \label{fig:row3}}
    \quad
\subfloat[]{%
    \includegraphics[width=0.23\linewidth]{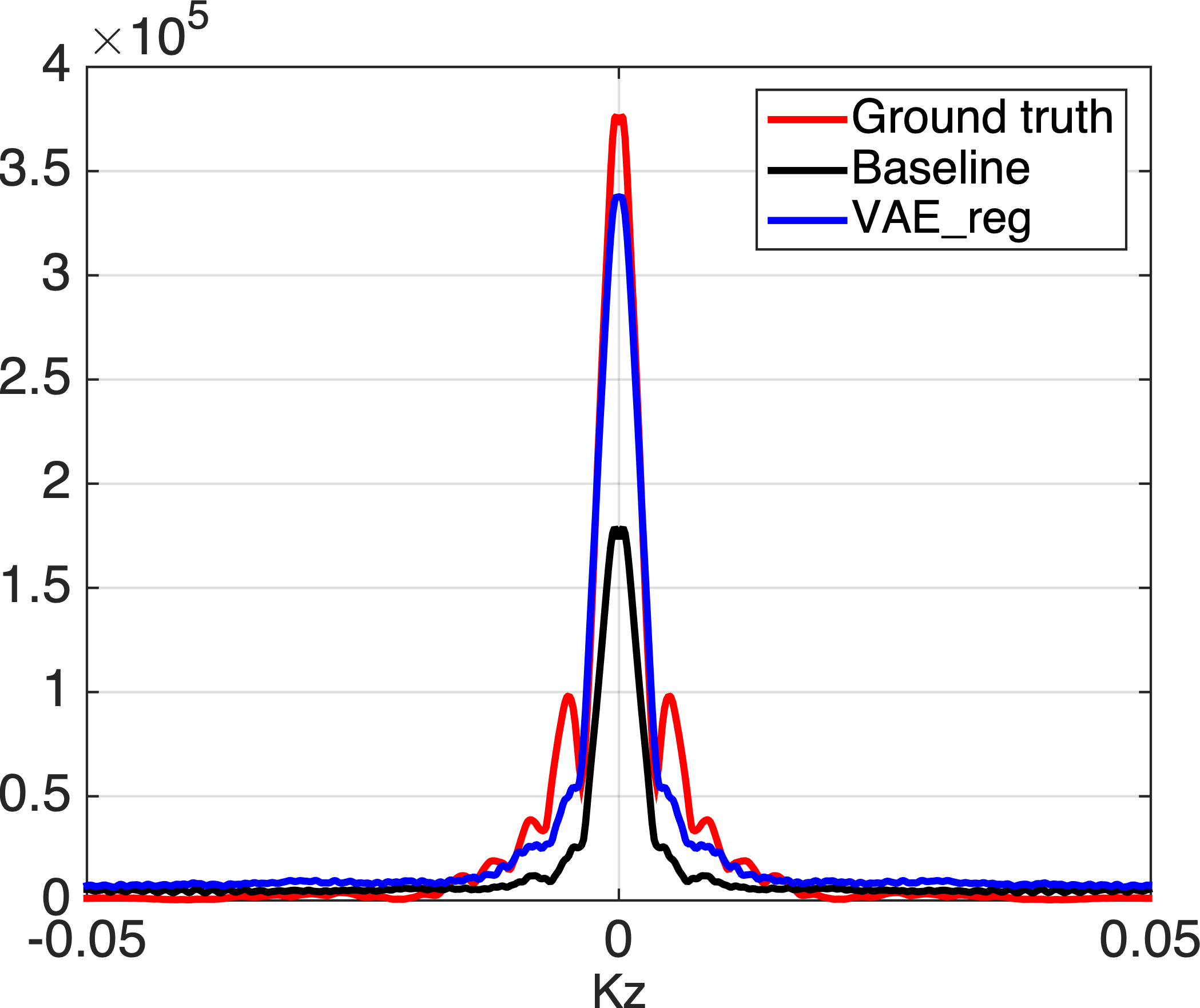}%
    \label{fig:row4}}}
\caption{Resolution analysis of the four imaging results as shown in Fig.~\ref{fig:smalltest}. The Kz spectra of our results (in blue) are much closer to those of the ground truth (in red) by comparing the baseline method (in black). That indicates that our imaging method yields higher spatial resolution than the baseline method.}
\label{fig:resolution}
\end{figure*}

Besides visualization of the resulting images, quantifying the spatial resolution will provide a different perspective to evaluate the quality of the results. Here, we employ the commonly used wavenumber analysis on our imaging results to help with justifying the quality of the resulting image resolution Our focus is on the velocity perturbation induced by the CO$_2$ leaks as shown in Fig.~\ref{fig:smalltest}. The perturbation can be obtained by subtracting the time-lapsed images from the baseline image (shown in Fig.~\ref{fig:baseline}), where the baseline image refers to the one without any leaks. Once the velocity perturbation is obtained after the subtraction, we employ the spatial Fourier transform to obtain the wavenumber~(i.e., the Kz spectrum), and we provide the plots (in Figs.~\ref{fig:row1} to \ref{fig:row4}) using all four imaging results shown in Fig.~\ref{fig:smalltest}. We observe that the Kz spectra of our results (in blue) are much closer to those of the ground truth (in red) by comparing the baseline method (in black) for all imaging results. That indicates that our imaging method yields higher spatial resolution than the baseline method. 

\begin{figure}[h]
\centerline{
\subfloat[]{%
    \includegraphics[width=0.75\linewidth]{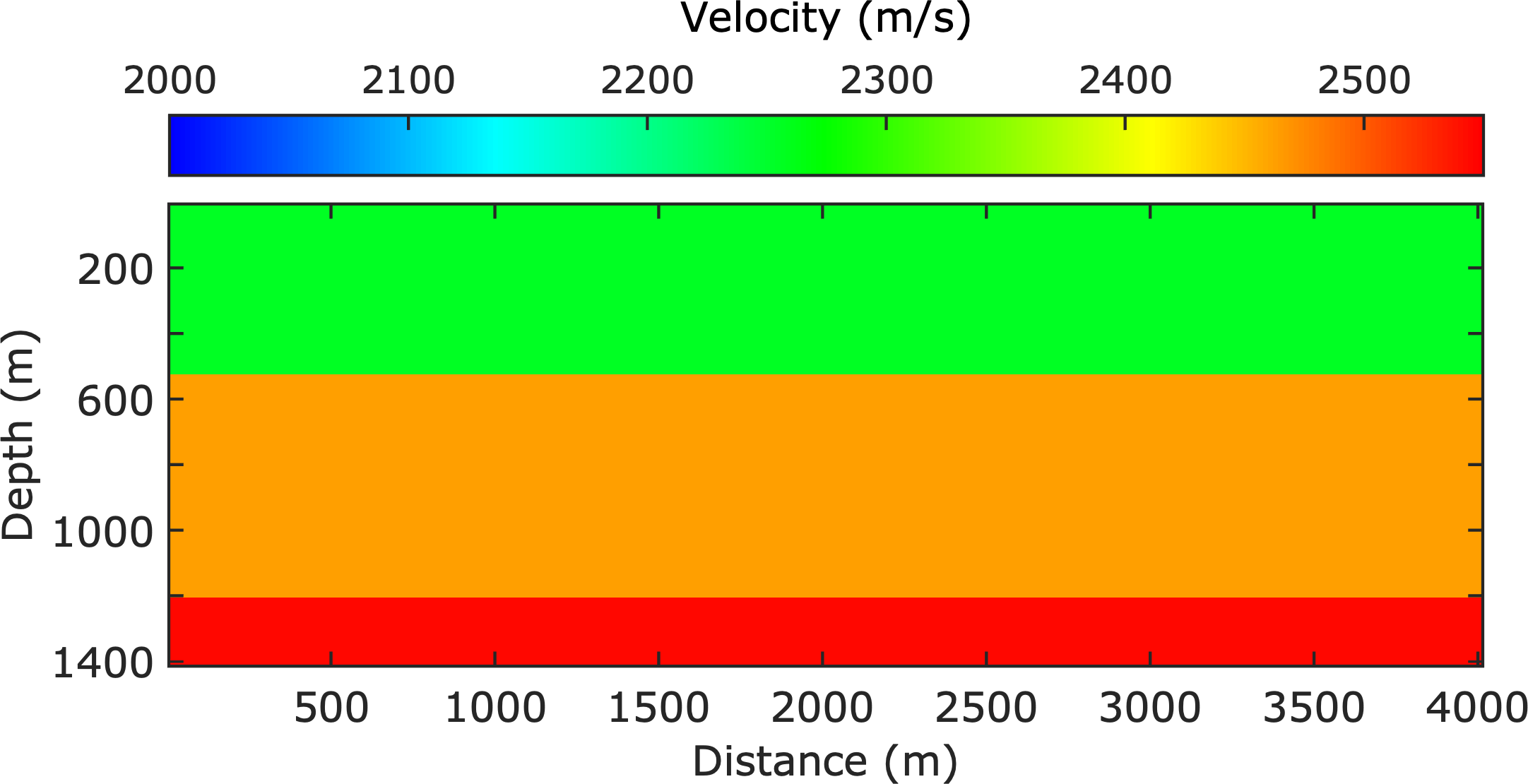}%
    \label{fig:baseline}}}
\caption{Illustration of the baseline velocity map.}
\label{fig:baseline}
\end{figure}

In this test, we study the performance of InversionNet using augmented data sets generated by our models on various leakage scenarios. Due to the lack of consideration of underlying physics knowledge, both autoencoder and vanilla VAE models do not help to improve the overall performance of InversionNet. On the other hand, our proposed model, VAE\_{reg}, is capable of generating physically realistic synthetic samples, which in turn will further improve in imaging all leakage cases. In particular, the imaging resolution on tiny leakage is significantly enhanced with the CO$_2$ plume much better resolved. It is worth mentioning that although the other proposed model, VAE\_{percep}, is capable of generating comparable synthetic samples to VAE\_{percep}, it may not lead to an improved overall imaging quality in this dataset. We still include this new model and its results since it may perform better in a different application and dataset.

Through the numerical tests and comparison, we conclude that the performance of InversionNet can be much improved in imaging all leakage cases. In particular, the imaging resolution on tiny leakage is significantly enhanced with the CO$_2$ plume much better resolved.

\subsection{Test 3: Determination of Augmented Data Size}

The size of the augmentation data is critical to the resulting performance. Without sufficient augmented data, the imaging results of InversionNet will be sub-optimal. In contrast, augmenting the original training set with too much data may lead to ``augmentation leak'', meaning that synthesized data dominate the training set and distorts the true data distribution~\cite{Zhao-2020-Image}. In this test, we aim to find out the best range of the amount of augmentation data for optimizing the performance of InversionNet. 

\begin{figure}[h]
    \centerline{\subfloat[]{\includegraphics[width=.5\linewidth]{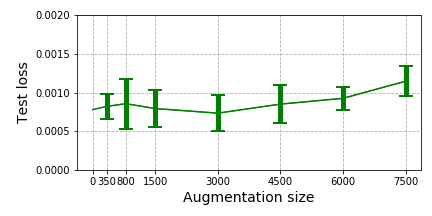}}
    \subfloat[]{\includegraphics[width=.5\linewidth]{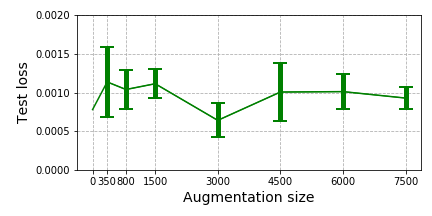}}}
    \caption{Mean and standard deviation of test loss for different augmentation sizes. Test loss of InversionNet using augmented data set generated with (a)  VAE with perception loss, (b) VAE with regularization. For all the tests, the same Small Leak data set~(see Table~\ref{tab:TestData}) are utilized as the test data.}
    \label{fig:augment_size}
\end{figure}

Through Tests~1 and 2, we learn that VAE with perception loss and VAE with regularization usually yield better results. Hence, we focus on the impact of the varying augmentation data size over those two models. We vary the size of the augmented data set among 350, 800, 1500, 3000, 4500, 6000, and 7500. For each case, we generate 5 different groups of augmentation data using our generative models. With all those groups of synthesized data being available, we train InversionNet with augmented data and test on the same Small Leak data as used in Test 2 (see Table~\ref{tab:TestData}). We report the corresponding loss values in terms of mean and standard deviation in Fig.~\ref{fig:augment_size}. We observe a general pattern ``decrease first $\longrightarrow$ bottom $\longrightarrow$ increase later'' from both results in Fig.~\ref{fig:augment_size}. This type of pattern has recently been discovered and analyzed in other data augmentation literature~\cite{Zhao-2020-Image, Karras-2020-Training}. It is mainly caused by an augmentation leak, which should be used as an indication to decide a reasonable augmentation data size. Interestingly, for both VAE with physics perception loss and VAE with  regularization, the smallest test loss values are achieved when 3,000 synthetic velocity maps are generated. Hence, throughout all the tests, we use 3,000 as the size of the augmented data set.

\subsection{Test 4: Hyper-parameter Selection} 
\label{test4}

Hyper-parameters play an important role in our generative models. In this test, we will study two critical hyper-parameters: the selection of the layers from the pre-trained VGG-19 network in Eq.~\eqref{eq:VAE_physicsloss} and the regularization parameter, $\gamma$, in Eq.~\eqref{eq:VAE_reg}. 

\begin{figure}[h]
    \centerline{\subfloat[]{\includegraphics[width=.5\linewidth]{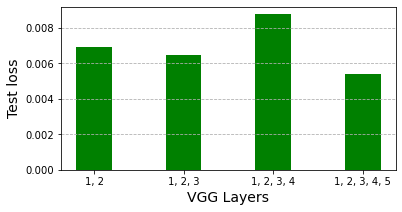}}
    \subfloat[]{\includegraphics[width=.5\linewidth]{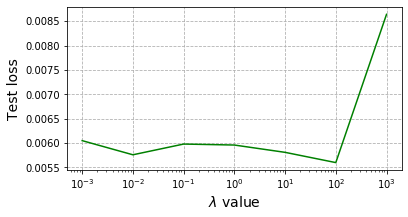}}}
    \caption{Visualization of the hyper-parameter versus loss values. (a) different combinations of layers selected from VGG-19 used in our VAE with perception loss. (b) Various values of the regularization parameter used in our VAE with regularization.}
    \label{fig:hyperparameter}
\end{figure}

To select the optimal VGG-19 layers for computing perception loss, there are 5 convolutional blocks in VGG-19 as shown in Fig.~\ref{fig:VAE_Style}(b). We follow a similar idea in~\cite{gatys2015neural} to select effective layers. Particularly, we choose from following four combinations of \textbf{(A)}~[conv1\_1, conv2\_1]; \textbf{(B)}~[conv1\_1, conv2\_1, conv3\_1]; \textbf{(C)}~[conv1\_1, conv2\_1, conv3\_1, conv4\_1]; and \textbf{(D)}~[conv1\_1, conv2\_1, conv3\_1, conv4\_1, conv5\_1]. We train VAE with perception loss using different combinations of layers, and compute test loss of each case on the same test dataset~(as shown in Fig.~\ref{fig:hyperparameter}(a)). We can observe that VAE with perception loss reaches smallest value when using combination \textbf{(D)}. That gives us the indication to use first layer of 5 convolutional blocks of VGG-19 to compute perception loss. 

Similarly, in order to select optimal $\lambda$, we choose 7 values evenly distributed on a log scale from $10^{-3}$ to $10^3$~(i.e. $10^{-3}$, $10^{-2}$, $10^{-1}$, $10^0$, $10^1$, $10^2$ and $10^3$). For each $\lambda$ value, we train a VAE with regularization, and test it on a common test dataset. The resulting test loss is shown in Fig.~\ref{fig:hyperparameter}(b). We observe that test loss is relative stable when $\lambda < 10^3$, and it reaches lowest value when $\lambda = 10^2$. So, $\lambda = 10^2$ becomes the optimal value for our problem.

\subsection{Additional Numerical Tests}
\label{additionalTests}

To help with understanding the performance of our proposed approaches, we carry out a few more numerical tests. Due to the page limit, we describe those additional tests in the Supplement to this paper. Those tests include: (1) a comparison of the performance with linear interpolation function; (2) test of our model using $\mathcal{L}_1$ and $\mathcal{L}_2$ loss functions; (3) a side-by-side comparison of our proposed VAE\_{reg} method with different adjacent time dynamics; and (4) the assessment of generated velocity maps and the convergence plots for training different models.

\section{Conclusion and Future Work}

\noindent \textbf{Conclusion.}~In this work, we develop several spatio-temporal data augmentation using convolutional neural networks to improve the data-driven seismic imaging in reconstructing ``rare events''. Our data augmentation techniques consider different physics information~(governing equations, observable perception, and physics phenomena), and further incorporate them through perception loss and regularization techniques. We evaluate the performance of our generative models to image very small CO$_2$ leaks from the subsurface using InversionNet. Through a thorough comparison and analysis, we show that physics information plays an important role in generating realistic and physically consistent synthetic data, which can be used to improve the representativeness of the training set. 

\noindent \textbf{Future Work.}~(1) The assessment of the synthesized data is a common challenge and an important question for any generative model including ours. Unlike some other disciplines ~(such as computer vision), certain physically-irrelevant metrics, such as Fr\'echet Inception Distance~(FID)~\cite{Heusel-2017-GANs}, can be used to evaluate the quality of the synthesized samples. A domain-specific metric will be needed to evaluate the resulting synthesized samples. This will be one of our future directions to explore. (2) In our problem of FWI, the data augmentation can be implemented from either the velocity domain or seismic domain. There are benefits to selecting either. We choose to augment in the velocity domain is to better leverage the prominent spatio-temporal dynamics, which is conceivable from the time-lapse velocity maps, and further incorporate them as part of the learning via different loss or additional regularization. In comparison, such spatio-temporal dynamics in the velocity implicitly embedded in the seismic wave after the forward modeling, which however becomes more challenging to capture. On the other hand, augmenting in the seismic has its benefit as well. It is a natural way and close to real-world problems, where we acquire seismic data without knowing what the subsurface structure is. We will study this in the near future. (3) The selection of hyper-parameters is important to the performance of our models. Currently, we employ a grid search strategy, i.e., dividing the domain of the hyper-parameter into a discrete grid. However, the performance of the grid search is justified by the training dataset, there could be some differences when applying our model to the test dataset. A more rigorous hyper-parameter selection strategy would benefit us in that perspective. There has been some discussion regarding this direction such as the evolutionary algorithm~\cite{young2015optimizing}. We will study this in the future. (4) Last but not least, we will study the applicability of our models to field datasets. In this work, we demonstrate the efficacy of our models on simulations, which would present similar physics to field data. Thus, it is reasonable to believe that our models would be applicable to real applications. However, some critical technical issues still needs to be carefully addressed, one of which is the model generalization ability. It is almost certain that there will be distribution drift when applying our models to different field datasets. To overcome this challenge, various techniques have been developed to improve the robustness and generalization ability of deep learning models under the context of scientific machine learning. That includes physics-informed deep neural networks~\cite{Physics-2021-Karniadakis}, semi-supervised learning~\cite{Survey-2021-Yang}, transfer learning~\cite{tan2018survey}, and many others. We will consider improving the generalization ability of our model in future R\&D efforts.

\section*{Acknowledgment}

This work was supported by the Center for Space and Earth Science at Los Alamos National Laboratory~(LANL), and by the Laboratory Directed Research and Development program under the project number 20210542MFR at LANL. We also thank two anonymous reviewers and the Associate Editor for their constructive suggestions and comments that improved the quality of this work.

\bibliographystyle{IEEEtran}
\bibliography{main}

\end{document}